\documentclass{article}

\usepackage{microtype}
\usepackage{graphicx}
\usepackage{subfigure}
\usepackage{booktabs} 

\usepackage{hyperref}

\usepackage[accepted]{icml2024}

\usepackage{amsmath}
\usepackage{amssymb}
\usepackage{mathtools}
\usepackage{amsthm}
\usepackage{enumitem}
\usepackage{amsfonts}
\usepackage{mathrsfs}
\usepackage{bm}
\usepackage{balance}
\usepackage{multirow}
\usepackage{booktabs}
\usepackage{wrapfig}

\usepackage[capitalize,noabbrev]{cleveref}

\newcommand{\norm}[1]{\left\lVert#1\right\rVert}

\theoremstyle{plain}
\newtheorem{theorem}{Theorem}[section]
\newtheorem{proposition}[theorem]{Proposition}

\theoremstyle{definition}
\newtheorem{definition}[theorem]{Definition}

\theoremstyle{remark}

\usepackage[textsize=tiny]{todonotes}

\icmltitlerunning{Verification of Machine Unlearning is Fragile}

\begin{document}

\twocolumn[
\icmltitle{Verification of Machine Unlearning is Fragile}



\icmlsetsymbol{equal}{*}

\begin{icmlauthorlist}
\icmlauthor{Binchi Zhang}{uva}
\icmlauthor{Zihan Chen}{uva}
\icmlauthor{Cong Shen}{uva}
\icmlauthor{Jundong Li}{uva}
\end{icmlauthorlist}

\icmlaffiliation{uva}{University of Virginia, Charlottesville, VA, USA}

\icmlcorrespondingauthor{Jundong Li}{jundong@virginia.edu}

\icmlkeywords{Machine Unlearning, Verifiable Machine Unlearning, Adversarial Unlearning}

\vskip 0.3in
]



\printAffiliationsAndNotice{}  

\begin{abstract}
As privacy concerns escalate in the realm of machine learning, data owners now have the option to utilize machine unlearning to remove their data from machine learning models, following recent legislation.
To enhance transparency in machine unlearning and avoid potential dishonesty by model providers, various verification strategies have been proposed. These strategies enable data owners to ascertain whether their target data has been effectively unlearned from the model.
However, our understanding of the safety issues of machine unlearning verification remains nascent.
In this paper, we explore the novel research question of whether model providers can circumvent verification strategies while retaining the information of data supposedly unlearned. 
Our investigation leads to a pessimistic answer: \textit{the verification of machine unlearning is fragile}.
Specifically, we categorize the current verification strategies regarding potential dishonesty among model providers into two types. 
Subsequently, we introduce two novel adversarial unlearning processes capable of circumventing both types. 
We validate the efficacy of our methods through theoretical analysis and empirical experiments using real-world datasets. 
This study highlights the vulnerabilities and limitations in machine unlearning verification, paving the way for further research into the safety of machine unlearning.
\end{abstract}

\section{Introduction}\label{sec:intro}
In the deep learning era, machine learning (ML) has grown increasingly data-dependent. 
A significant volume of personal data has been utilized to train real-world ML systems.
While the plentiful use of personal data has facilitated the advancement of machine learning, it simultaneously poses a threat to user privacy and has resulted in severe data breaches~\citep{nguyen2022survey}.
Consequently, recent regulations and laws~\citep{GDPR,CCPA} have mandated a novel form of request: the elimination of personal data and its impact from a trained ML model, known as machine unlearning~\citep{cao2015towards} (MUL).
In recent years, a proliferation of MUL techniques has emerged, employing various methods to remove certain training data~\citep{nguyen2022survey,xu2023machine}.

Despite the success achieved, current MUL techniques are still a black box for data owners, i.e., data owners cannot monitor the unlearning process and ascertain whether their data has been truly unlearned from the model~\citep{eisenhofer2022verifiable,xu2023machine}. 
For instance, many technology companies, e.g., Google and Microsoft, provide machine learning as a service (MLaaS).
Under MLaaS, individual data owners upload personal data to the server. 
The server then trains an ML model using the collected dataset and provides its predictive functionality as a service to the data owners~\citep{sommer2022athena}. 
In particular, data owners might want their data to be unlearned as long as this service is no longer required.
However, after sending an unlearning request, data owners will not receive any proof that their data was indeed unlearned.
Without the ability to verify, data owners have to trust the model provider blindly for the efficacy and integrity of the unlearning process.
On the other hand, the model provider might deceive data owners and pretend to have their data unlearned to avoid potential deterioration of the model utility and extra computational cost caused by unlearning~\citep{eisenhofer2022verifiable}.
Other than MLaaS, similar problems also exist in other domains involving personal data, e.g., social media and financial systems.
To address the above issue, recent studies have started exploring the verification strategy for MUL techniques from different aspects, e.g., injecting backdoor data into the training set or reproducing the unlearning operations.
Regarding the model provider's potential dishonesty, a natural question arises: are current MUL verification strategies ensured to be safe?
Specifically, we aim to study the following research question:
\begin{center}
    \textit{Can model providers successfully circumvent current verification strategies with dishonest unlearning?}
\end{center}

To answer this question, we systematically investigate the vulnerability of verification methods and obtain a pessimistic answer: \textit{current MUL verification is fragile}, i.e., data owners may fail to verify the integrity of the unlearning process using current MUL verification strategies. 
To support our study, we first categorize existing verification strategies into two types: \emph{backdoor verification} and \emph{reproducing verification}. 
We then propose an adversarial unlearning process that can successfully circumvent both of them, i.e., \textit{always satisfies these two types of verification} but \textit{still preserves the information of unlearned data}.
During unlearning, our approach selects the mini-batches excluding the unlearned data to effectively evade the detection by existing verification strategies, even with the most stringent reproducing verification strategy. 
Meanwhile, the selected batches are designed to yield model updates akin to those that would be produced by the unlearned data.
By deliberately choosing the retained data that mimics the influence of the unlearned data in training, the unlearned model yielded by our adversarial method retains the information from the unlearned data.
In addition, we propose a \textit{weaker but more efficient} adversarial unlearning process that can deceive a subset of reproducing verification by forging the unlearning processes directly from the original training steps.
We provide theoretical guarantees for the efficacy of our approaches. 
Furthermore, we conduct comprehensive empirical experiments to validate the efficacy of our proposed adversarial unlearning processes on real-world datasets.
We highlight our main contributions as follows:
\begin{itemize}[leftmargin=*]
\item We propose two adversarial unlearning methods that can circumvent both types of current MUL verification strategies (backdoor and reproducing) while preserving the information of unlearned data. 
\item We prove the capacity of the proposed adversarial methods to satisfy the stringent reproducing verification. 
We also prove that they can preserve the unlearned model utility as the original training and improve the efficiency.
\item We conduct empirical experiments to verify the efficacy of our proposed adversarial methods in real-world datasets, exposing the vulnerability of MUL verification.
\end{itemize}

\section{Related Works}
\subsection{Machine Unlearning}
The goal of MUL is to make a trained ML model forget some specific training data~\citep{cao2015towards}.
A simple but powerful way of unlearning is to retrain the model from scratch, which is also called exact unlearning~\citep{cao2015towards,bourtoule2021machine,kim2022efficient,chen2022graph}.
In exact unlearning, the unlearned model is ensured to behave as has never seen the unlearned data, which satisfies the goal of MUL.
Following this path, \citeauthor{cao2015towards} first proposed the retraining-based methodology for MUL. Follow-up studies~\citep{bourtoule2021machine,kim2022efficient,chen2022graph} took steps to improve the efficiency of the retraining framework on the image and graph data.
Despite the efforts to improve efficiency, the retraining-based framework still has difficulty accommodating frequent unlearning requests in the real world~\citep{chien2023efficient}.
Due to the large computational cost of exact unlearning, approximate unlearning was proposed that efficiently updates the original model to estimate the retrained model~\citep{guo2020certified,ullah2021machine,izzo2021approximate,zhang2022prompt,pan2023unlearning,wu2023gif,wu2023certified,che2023fast,warnecke2023machine,zhangtowards,dong2024idea}.
A common way to update the original model is to use the influence function~\citep{koh2017understanding}, which can be seen as conducting a single Newton step~\citep{boyd2004convex,sekhari2021remember,neel2021descent} to the model.
In particular, approximate unlearning can be certified if the distance between the unlearned model and the retrained model is bounded in the probability space~\citep{nguyen2022survey}.

\subsection{Verification for Machine Unlearning}

\textbf{Backdoor Verification.}
Backdoor verification of MUL requires data owners to actively inject backdoor poisoned data (e.g., changing the original label to a different one for misleading) as backdoor triggers~\citep{sommer2022athena,gao2022verifi,guo2023verifying}.
In this way, the model trained on the backdoor data can misclassify the backdoor data as the modified classes.
To verify the integrity of unlearning, data owners can deliberately request the model provider to unlearn the backdoor data.
If the model provider honestly unlearns the backdoor data, the predictions are supposed to be the original label with high confidence (backdoor data is not triggered);
otherwise, the predictions can still be misled to the poisoned label (backdoor data is triggered).
In addition, \citeauthor{sommer2022athena} formulated the backdoor verification as a hypothesis test, enabling a probabilistic guarantee of a successful verification.

\textbf{Reproducing Verification.}
Inspired by verifiable computation techniques, reproducing verification of MUL requires the model provider to provide a \textit{proof of unlearning (PoUL)} that records the operations for unlearning, and data owners can reproduce every unlearning step to verify the integrity of unlearning.
Considering that the model provider might deceive the data owner while conducting exact unlearning, \citeauthor{thudi2022necessity} first introduced the proof of learning (PoL) technique to verify the model retraining operation. 
Although they showed that the PoL can be forged by model providers, their forging strategy is not realistic and brings limitations to understanding the safety of MUL verification.
Following this paradigm, \citeauthor{weng2022proof} presented a trusted hardware-empowered PoUL technique with SGX enclave~\citep{costan2016intel}. With the trusted hardware, their framework provides a better safety guarantee for verifying MUL.
Recently, \citeauthor{eisenhofer2022verifiable} proposed the first cryptographic definition of verifiable unlearning and instantiated the PoUL with SNARKs~\citep{costan2016intel} and hash chains. Their verification framework has a safety guarantee from a cryptographic perspective.

Comparing different types of verification, backdoor verification requires data owners to inject poisoned data beforehand, while reproducing verification requires model providers to generate proof for the unlearning operations. 
In addition, \citeauthor{xu2023machine} summarized the current MUL evaluation methods (mainly for approximate unlearning), e.g., accuracy~\citep{golatkar2020eternal,golatkar2021mixed,mehta2022deep}, relearning time~\citep{kim2022efficient,chundawat2023zero,tarun2023deep}, and membership inference attack~\citep{chen2021machine} as verification strategies.
However, even knowing the evaluation results, data owners still need to compare the results with the model retrained from scratch (exact unlearning method), which is unknown to data owners in practice.
More importantly, evaluation methods fail to take into account the dishonest behaviors of model providers. 
Therefore, we do not focus on the safety problem of evaluation in this paper.

\begin{figure}
    \centering
    \includegraphics[clip, trim=3cm 5.5cm 3cm 5.5cm, width=\linewidth]{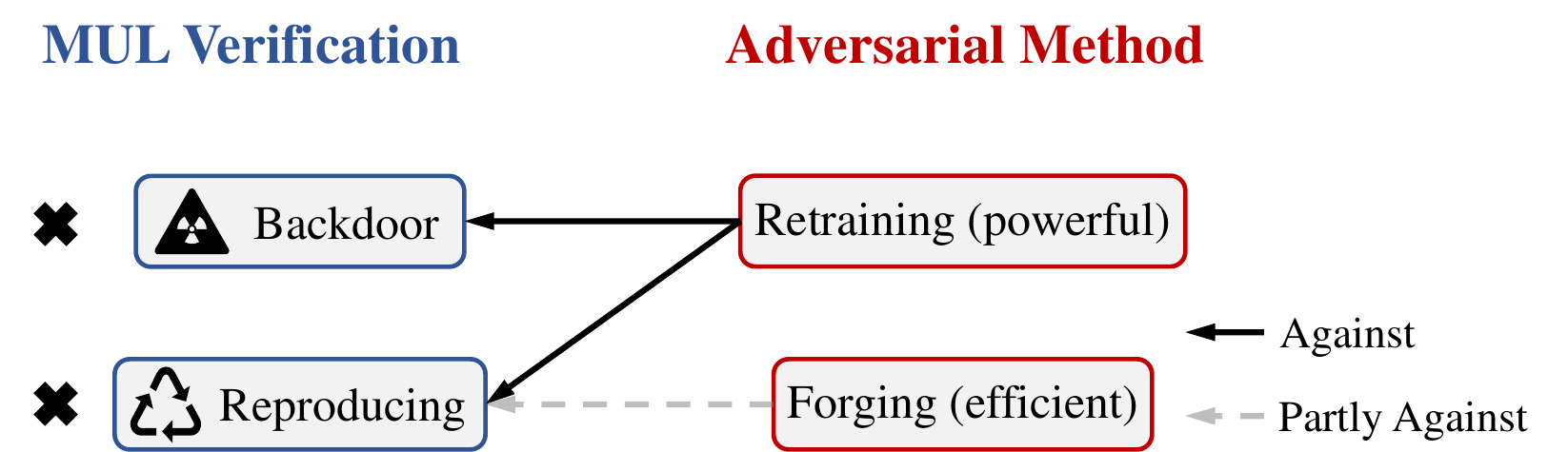}
    \vspace{-1.5em}
    \caption{The connection of our threat model and different verification strategies. 
    Our retraining method can deceive the backdoor and reproducing verification, and our forging method can only deceive a subset of reproducing verification but with better efficiency.} 
    \label{fig:connection}
    \vspace{-12pt}
\end{figure}

\section{Threat Model}
In our problem, we consider the data owner as a victim and the model provider as an adversary regarding safety in MUL verification. 
First, the model provider trains an ML model based on a training dataset provided by the data owners.
After the data owners send an unlearning request, the model provider deceives the data owners that their data has been unlearned, while the model provider updates the model with an adversarial unlearning process rather than normal unlearning methods to preserve the information of the unlearned data.
Next, we introduce the settings of our adversarial unlearning process from two perspectives: the adversary’s goal and knowledge.

\paragraph{Adversary's Goal.}
The goal of using an adversarial unlearning process is to preserve the information of unlearned data during unlearning while satisfying the reproducing and backdoor verification.
Note that the model provider might cheat the data owner for two benefits, i.e., \textit{better model utility} and \textit{lower computational cost}.
Hence, other than preserving the unlearned data, the adversary's goal should also include these benefits.

\paragraph{Adversary's Knowledge.}
The model owner has complete access to the training process, e.g., training data, unlearned data, and the model.

As a tentative step toward exploring the vulnerability of MUL verification, we mainly focus on the verification of \textbf{exact unlearning}, i.e., verifying whether the model provider honestly retrains the model from scratch.
We conclude the connection between our threat model and different types of verification strategies for MUL in \cref{fig:connection}.

\section{Methodology}
In this section, we introduce our design of an adversarial unlearning process that deceives both the reproducing verification and the backdoor verification.
To satisfy the reproducing verification, the model provider should provide a valid Proof of Retraining (PoRT)\footnote{For better clarity, we use the term PoRT to replace the aforementioned PoUL as they are the same under exact unlearning.}.
Hence, we first introduce the notations and background knowledge of PoRT before proposing our adversarial unlearning process.


\subsection{Preliminary}
\paragraph{Notation.}
We denote $\mathcal{D}$ as a training dataset with $n$ data samples and $f_{\bm{w}}$ as an ML model with $\bm{w}$ collecting the learnable parameters.
Let $\mathcal{A}$ be a learning process that takes the training set $\mathcal{D}$ as input and outputs the optimal $\bm{w}^*$ that minimizes the empirical risk on $\mathcal{D}$.
In particular, a learning process $\mathcal{A}$ is used to minimize the empirical loss function 
\begin{equation}
\mathcal{A}(\mathcal{D})=\mathrm{argmin}_{\bm{w}}\mathcal{L}(\bm{w},\mathcal{D}),
\end{equation}
where $\mathcal{L}(\bm{w},\mathcal{D})=\frac{1}{|\mathcal{D}|}\sum_{(\bm{x},y)\in\mathcal{D}}l(f_{\bm{w}}(\bm{x}),y)$ is the empirical risk over $\mathcal{D}$, $(\bm{x},y)$ denotes the pair of the input data and the output label, and $l(\cdot)$ denotes a task-specific loss function, e.g., cross-entropy loss.
We use $\mathcal{D}_u$ to denote the set of data to be unlearned. 
Consequently, the model retrained from scratch can be obtained by $\mathcal{A}(\mathcal{D}\backslash\mathcal{D}_u)$.

\paragraph{Proof of Retraining.}
In reproducing verification, the model provider is required to provide a PoRT, and the data owner or a third-party verifier can reproduce all the retraining operations in the PoRT to verify the integrity of unlearning.
A prevalent instantiation of PoRT is to record the trajectory of retraining during the unlearning process~\citep{thudi2022necessity,weng2022proof,eisenhofer2022verifiable}, denoted by $\mathcal{P}_r=\{\bm{w}_r^{(t)}, d_r^{(t)}, g_r^{(t)}\}_{t\in I}$ where $\bm{w}_r^{(t)}$ denotes the intermediate model parameter during retraining and $d_r^{(t)}$ denotes the data used for deriving $\bm{w}_r^{(t)}$. 
$g_r^{(t)}$ denotes the updating function and $I$ is the set of indices to the intermediate learning steps during retraining ($I=\{1,2,\dots, T\}$ where $T$ is the number of model updating steps). 
In addition, a Proof of Training (PoT) can be defined similarly as the PoRT, i.e., $\mathcal{P}_t=\{\bm{w}^{(t)}, d^{(t)}, g^{(t)}\}_{t\in I}$, while the difference is that the unlearned data should be involved in the training but not in the retraining.
In this paper, we assume that the equivalence of adopted models and the models appear on the PoT ($\bm{w}^{(T)}$) and the PoRT ($\bm{w}_r^{(T)}$) can be verified, otherwise, reproducing verification will never be possible and the problem becomes trivial.

\paragraph{Reproducing Verification.}
We next provide a formal definition of reproducing verification~\citep{thudi2022necessity,weng2022proof,eisenhofer2022verifiable}.
As the retraining process can be divided into iterative steps, the PoRT can be seen as an ordered set of triplets $\{\bm{w}_r^{(t)},d_r^{(t)},g_r^{(t)}\}$.
Each triplet describes an iterative operation that updates the retrained model based on an updating function as $\bm{w}_r^{(t)}=g_r^{(t)}(\bm{w}_r^{(t-1)},d_r^{(t)})$.
As an example, we can instantiate the updating function as the commonly used mini-batch stochastic gradient descent~\citep{goodfellow2016deep}:
\begin{equation}\label{eq:minibatch_gd}
\small
g_r^{(t)}(\bm{w}_r^{(t-1)},d_r^{(t)})=\bm{w}_r^{(t-1)}-\gamma^{(t)}\nabla\mathcal{L}(\bm{w}_r^{(t-1)},d_r^{(t)}),
\end{equation}
where $\gamma^{(t)}$ denotes the learning rate.
We can define a valid PoRT that can satisfy the reproducing verification as follows.
\begin{definition}\label{def:valid_proof}
A valid Proof of Retraining is defined as $\mathcal{P}_r=\{\bm{w}_r^{(t)}, d_r^{(t)}, g_r^{(t)}\}_{t\in I}$ that satisfies the following two properties:
\begin{enumerate}[leftmargin=*]
\item Reproducible: $\forall$ $t\in I$, $\|\bm{w}_r^{(t)}-g_r^{(t)}(\bm{w}_r^{(t-1)},d_r^{(t)})\|\leq\varepsilon$;
\item Removable: $\forall$ $t\in I$, $d_r^{(t)}\cap\mathcal{D}_u=\emptyset$.
\end{enumerate}
\end{definition}
In the reproducible property, \citeauthor{thudi2022necessity} set $\varepsilon$ as a threshold for error tolerance, allowing the verifier to consider some numerical imprecision when reproducing the update rule, and $\|\bm{w}_r^{(t)}-g_r^{(t)}(\bm{w}_r^{(t-1)},d_r^{(t)})\|$ is called the verification error at step $t$. 
In \citep{weng2022proof,eisenhofer2022verifiable}, the threshold $\varepsilon$ can be reduced to an exact $0$ with the help of SNARK~\citep{setty2020spartan} and Intel SGX~\citep{costan2016intel}.
Hence, we consider both the cases of $0$ and $\varepsilon$ thresholds.

\begin{figure}
    \centering
    \includegraphics[clip, trim=6cm 5cm 6cm 5cm, width=\linewidth]{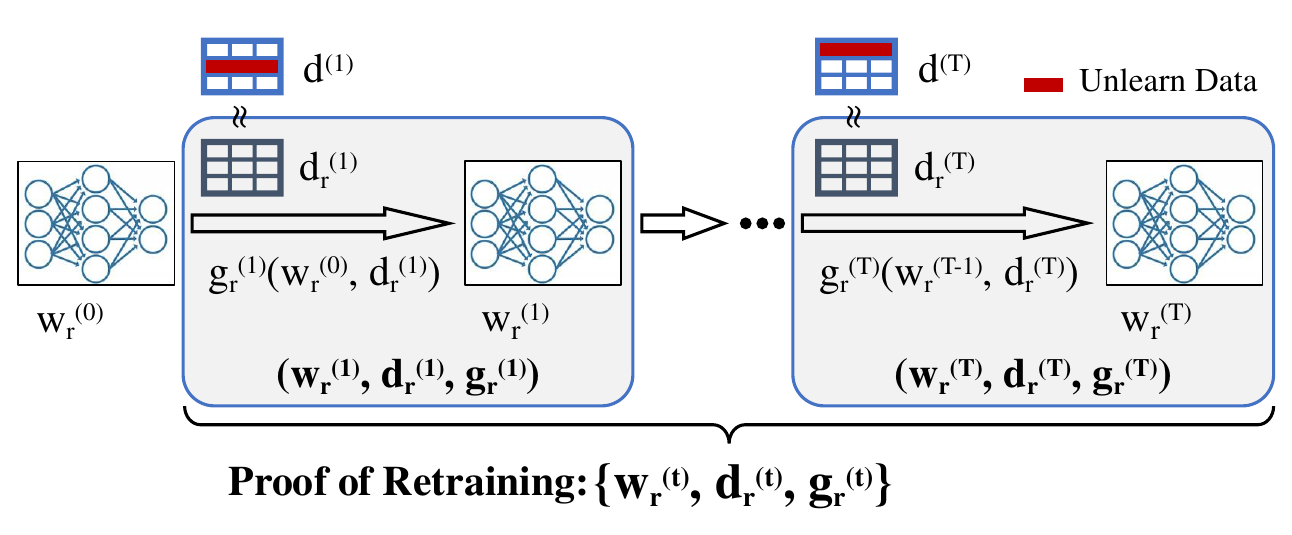}
    \vspace{-2em}
    \caption{An illustration of the retraining-based adversarial unlearning framework. The PoRT is generated based on the retraining process where the mini-batch $d_r^{(t)}\in\mathcal{D}\backslash\mathcal{D}_u$ sampling is guided by the similarity with $d^{(t)}\in\mathcal{D}$ in gradient.}
    \label{fig:retrain}
    \vspace{-10pt}
\end{figure}

\subsection{First Adversarial Method (Retraining)}\label{sec:retrain based}
We first introduce our adversarial unlearning process against both reproducing and backdoor verification.
Recall that the adversary's goal is to satisfy the unlearning verification while preserving the information of unlearned data.
To satisfy the reproducing verification with a $0$ verification error for each updating function, the model provider has to follow exactly the retraining process to generate the PoRT, strictly ensuring the reproducible and removable properties.
However, inspired by recent studies of the possibility that different training data might lead to a similar gradient descent step~\citep{shumailov2021manipulating,thudi2022necessity}, we can find data in the retained set $\mathcal{D}\backslash\mathcal{D}_u$ that can yield a similar gradient as the unlearned data.
In particular, this idea is based on the commonly used mini-batch gradient descent.
Under full-batch gradient descent, the gradients in the original training and the retraining are computed over the whole $\mathcal{D}$ and $\mathcal{D}\backslash\mathcal{D}_u$.
Their difference is deterministic and cannot be manipulated.
In contrast, under mini-batch gradient descent, the gradients in the original training and the retraining are computed over \textit{random samples} in $\mathcal{D}$ and $\mathcal{D}\backslash\mathcal{D}_u$.
With the randomness, it is possible to deliberately select the samples that induce a similar model update as the mini-batches in $\mathcal{D}$ from $\mathcal{D}\backslash\mathcal{D}_u$ in the retraining.
In this way, the retrained model might still learn from the unlearned data as the original training, even without explicitly using them.

As discussed above, we convert the problem of the adversarial unlearning process into selecting $d_r^{(t)}\in\mathcal{D}\backslash\mathcal{D}_u$ that yields similar model updates as $d^{(t)}\in\mathcal{D}$.
It is evident that if the original mini-batch contains no unlearned data $d^{(t)}\cap\mathcal{D}_u=\emptyset$, we can directly set $d_r^{(t)}=d^{(t)}$.
If the original mini-batch contains unlearned data, then our goal is to find $d_r^{(t)}\in\mathcal{D}\backslash\mathcal{D}_u$ that makes $\nabla\mathcal{L}(\bm{w}_r^{(t-1)},d^{(t)})$ and $\nabla\mathcal{L}(\bm{w}_r^{(t-1)},d_r^{(t)})$ as close as possible.
Regarding this goal, previous studies simply leveraged random sampling multiple times and chose the one yielding the smallest distance~\citep{shumailov2021manipulating,thudi2022necessity}.
\begin{equation}\label{equ:random}
\small
d_r^{(t)}=\mathrm{argmin}_{d_i}\|\nabla\mathcal{L}(\bm{w}_r^{(t-1)},d^{(t)})-\nabla\mathcal{L}(\bm{w}_r^{(t-1)},d_i)\|,
\end{equation}
where $i=1,\dots,s$, $d_1,\dots,d_s\sim\mathcal{D}\backslash\mathcal{D}_u$, and we denote the right-hand side of \cref{equ:random} as $\mathcal{S}_r(\bm{w}_r^{(t-1)};d^{(t)})$.
However, this method is computationally expensive and does not provide theoretical guarantees.
In this paper, we replace the unlearned data in the original batch: $(\bm{x}_u,y_u)\in d^{(t)}\cap\mathcal{D}_u$ with its (class-wise) closest neighbor in the retained data: 
\begin{equation}
\mathcal{N}(\bm{x}_u,y_u)=\mathrm{argmin}_{(\bm{x},y)\in\mathcal{D}\backslash\mathcal{D}_u,y=y_u}\|\bm{x}-\bm{x}_u\|.
\end{equation}
Consequently, our proposed mini-batch selection method can be expressed as
\begin{equation}\label{eq:nearest neighbor}
\small
\mathcal{S}_n(d^{(t)})=(d^{(t)}\backslash\mathcal{D}_u)\cup\{\mathcal{N}(\bm{x},y)|(\bm{x},y)\in d^{(t)}\cap\mathcal{D}_u\}.
\end{equation}
It is worth noting that we observe that using $\mathcal{S}_r$ can be better than $\mathcal{S}_n$ sometimes in practice, i.e., $\nabla\mathcal{L}(\bm{w}_r^{(t-1)},\mathcal{S}_r(\bm{w}_r^{(t-1)};d^{(t)}))$ might be closer to $\nabla\mathcal{L}(\bm{w}_r^{(t-1)},d^{(t)})$ than $\nabla\mathcal{L}(\bm{w}_r^{(t-1)},\mathcal{S}_n(\bm{w}_r^{(t-1)};d^{(t)}))$, as long as setting a large enough sample size $s$.
However, our proposed closest neighbor selection is still necessary as it provides a worst-case upper bound for theoretical analysis and a more efficient way of adversarial unlearning when considering real-world threats.
Consequently, we can generate a PoRT based on our adversarial unlearning process with carefully selected mini-batches.
We provide a clear illustration of our adversarial unlearning framework in \cref{fig:retrain}.
In addition, the detailed algorithm is shown in~\cref{alg:adv1}.
Specifically, the following properties can be guaranteed for \cref{alg:adv1}.

\begin{algorithm}[t]
   \caption{Retraining-based Adversarial Unlearning Algorithm}
   \label{alg:adv1}
\begin{algorithmic}
   \STATE {\bfseries Input:} Training data $\mathcal{D}$, unlearned data $\mathcal{D}_u$.
   \STATE {\bfseries Output:} Proof of Retraining $\mathcal{P}_r$.
   \STATE Initialize $\bm{w}_r^{(0)}$ and $\mathcal{P}_r\leftarrow\emptyset$.
   \FOR{$t=1$ {\bfseries to} $T$}
   \STATE Uniform mini-batch sampling $d^{(t)}\in\mathcal{D}$.
   \STATE Choose $d_r^{(t)}\leftarrow\mathcal{S}_r(\bm{w}_r^{(t-1)};d^{(t)})$ or $d_r^{(t)}\leftarrow\mathcal{S}_n(d^{(t)})$.
   \STATE $\bm{w}_r^{(t)}\leftarrow g_r^{(t)}(\bm{w}_r^{(t-1)},d_r^{(t)})$.
   \STATE $\mathcal{P}_r\leftarrow\mathcal{P}_r\cup(\bm{w}_r^{(t)},d_r^{(t)},g_r^{(t)})$.
   \ENDFOR
\end{algorithmic}
\end{algorithm}

\begin{proposition}\label{pro:valid}
\cref{alg:adv1} returns a valid Proof of Retraining under the threshold $\varepsilon=0$.
\end{proposition}


\begin{proposition}\label{pro:cvg}
Let $C_D:=\max_c\max_{\bm{x}\in c}\min_{\bm{z}\in c}\|\bm{x}-\bm{z}\|$, where $c$ denotes the class in the label domain.
For the continuity of loss functions, we assume (i). $\|\nabla_{\bm{w}}l(\bm{w},\bm{x})\|\leq G$, (ii). $\|\frac{\partial^2l(\bm{w},\bm{x})}{\partial\bm{w}\partial\bm{x}}\|\leq L_x$, and (iii). $\|\nabla^2_{\bm{w}}l(\bm{w},\bm{x})\|\leq L$.
Let $\gamma^{(t)}=\gamma\leq\frac{1}{L}$ and $m$ be the size of mini-batches, for \cref{alg:adv1}, we have
\begin{equation*}
\small
\begin{aligned}
\mathrm{E}_T[\|\nabla\mathcal{L}(\bm{w}_r^{(T)},\mathcal{D})\|^2]&\leq\frac{\mathcal{L}(\bm{w}_r^{(0)},\mathcal{D})-\mathcal{L}(\bm{w}_r^{(T)},\mathcal{D})}{\gamma T} \\
&+\frac{\gamma L}{2}(G^2+p_uB^2)+(1-\gamma L)p_uGB,
\end{aligned}
\end{equation*}
where $p_u=1-(1-\frac{|\mathcal{D}_u|}{|\mathcal{D}|})^m$ and $B=L_xC_D$.
\end{proposition}
Assumptions (i) and (iii) are the same as in \citep{ajalloeian2020convergence}, indicating that $l(\cdot,\bm{x})$ is $L$-smooth and $G$-Lipschitz continuous. In addition, Assumption (ii) is the same as in \citep{thudi2022necessity}, indicating that $\nabla_{\bm{w}}l(\bm{w},\cdot)$ is $L_x$-Lipschitz. 
We provide the proof of \cref{pro:valid} and \cref{pro:cvg} in \cref{sec:proof}.
\cref{pro:valid} ensures that \cref{alg:adv1} can satisfy the reproducing verification;
\cref{pro:cvg} guarantees that the retraining process can reach a neighborhood of the optimum over $\mathcal{D}$, i.e., the adversarial unlearning process can still make the retrained model learn from $\mathcal{D}_u\subset\mathcal{D}$.
Next, regarding the backdoor verification, if the retrained model reaches the exact optimum over $\mathcal{D}$, the backdoor poisoned data in the unlearned set can still be triggered, and the adversarial unlearning process will be recognized.
However, according to \cref{pro:cvg}, the radius of the neighborhood $\frac{L\gamma}{2}(G^2+p_uB^2)+(1-\gamma L)p_uGB$ (the distance between the retrained model and the original optimum) will increase after injecting backdoor data, which reduces the probability of backdoor data being triggered.
The rationale is that after flipping the label of the backdoor data $\bm{x}_b$ from $y_b$ to $y_b^\prime$, the value of $\min_{\bm{x}\in c_{y_b^\prime}}\|\bm{x}-\bm{x}_b\|$ can increase because $\bm{x}_b$ is actually in the class $y_b$ with a different distribution.
Instead of learning the mapping of backdoor data $\bm{x}_b\rightarrow y_b^\prime$, our adversarial unlearning process uses the data truly from the class $y_b^\prime$ to replace each of the backdoor data.
Hence, our unlearned model cannot be triggered by the backdoor poisoned data.

\subsection{Second Adversarial Method (Forging)}
Although our first adversarial unlearning method can deceive the verification of MUL while preserving the information of unlearned data, the computational overhead of our method is not better than naive retraining, limiting the benefits of the model provider earned from the adversarial unlearning process.
Hence, we propose another adversarial unlearning process that is more computationally efficient.
As a trade-off, the power of our second adversarial method becomes weaker, i.e., it can only deceive the reproducing verification under an $\varepsilon>0$ threshold.
Inspired by \citep{thudi2022necessity}, if the model provider records a PoT during the original training, the computation of generating a PoRT can be reduced by reusing the PoT.
In~\citep{thudi2022necessity}, the model provider can update the PoT using a forging map and obtain a valid PoRT under the $\varepsilon$-threshold reproducing verification.
In particular, the forging map is to replace the overlapping batches $d^{(t)}\cap\mathcal{D}_u\neq\emptyset$ with a batch $d_r^{(t)}\in\mathcal{D}\backslash\mathcal{D}_u$ that induces a similar model update as $d^{(t)}$ in each triplet $\{\bm{w}^{(t)},d^{(t)},g^{(t)}\}$, i.e., $\nabla\mathcal{L}(\bm{w}^{(t-1)},d^{(t)})\approx\nabla\mathcal{L}(\bm{w}^{(t-1)},d_r^{(t)})$.
However, the forging map is not realistic because it preserves the model parameters unchanged $\bm{w}_r^{(t)}=\bm{w}^{(t)}$, which can be easily recognized by the verifier (under multiple verification requests, the verifier will receive PoRTs with the same model parameters in each time).
In this paper, we propose a novel forging map $\mathcal{F}:\mathcal{P}_t\rightarrow\mathcal{P}_r$ that \textit{directly updates both \underline{the model parameters} and \underline{the batch data} of each triplet in the PoT to generate a valid PoRT instead of retraining}.
We formulate our proposed forging map as a triplet-wise updating function, i.e., for any $t\in\mathcal{I}$, 
\begin{equation}\label{eq:forging_map}
(\bm{w}_r^{(t)},d_r^{(t)},g_r^{(t)})=\mathcal{F}^{(t)}(\bm{w}^{(t)},d^{(t)},g^{(t)}).
\end{equation}
In this way, each triplet in $\mathcal{P}_r$ can be generated separately based on the corresponding triplet in $\mathcal{P}_t$.
Normally, the model updating function $g$ remains unchanged during the retraining process, and we mainly focus on updating $\bm{w}^{(t)}$ and $d^{(t)}$ with $\mathcal{F}^{(t)}$.
We divide the updating of $\bm{w}^{(t)}$ and $d^{(t)}$ into two cases: excluding unlearned data $\mathcal{I}_e=\{t\,|\,d^{(t)}\cap\mathcal{D}_u=\emptyset\}$ and including unlearned data $\mathcal{I}_n=\{t\,|\,d^{(t)}\cap\mathcal{D}_u\neq\emptyset\}$ ($\mathcal{I}_e$ and $\mathcal{I}_n$ are used to denote the index set of the triplets in different cases).
We assume that the number of unlearned data is far less than the retained data. Next, we discuss the two cases separately.

\begin{figure}
    \centering
    \includegraphics[clip, trim=6cm 2cm 6cm 2cm, width=\linewidth]{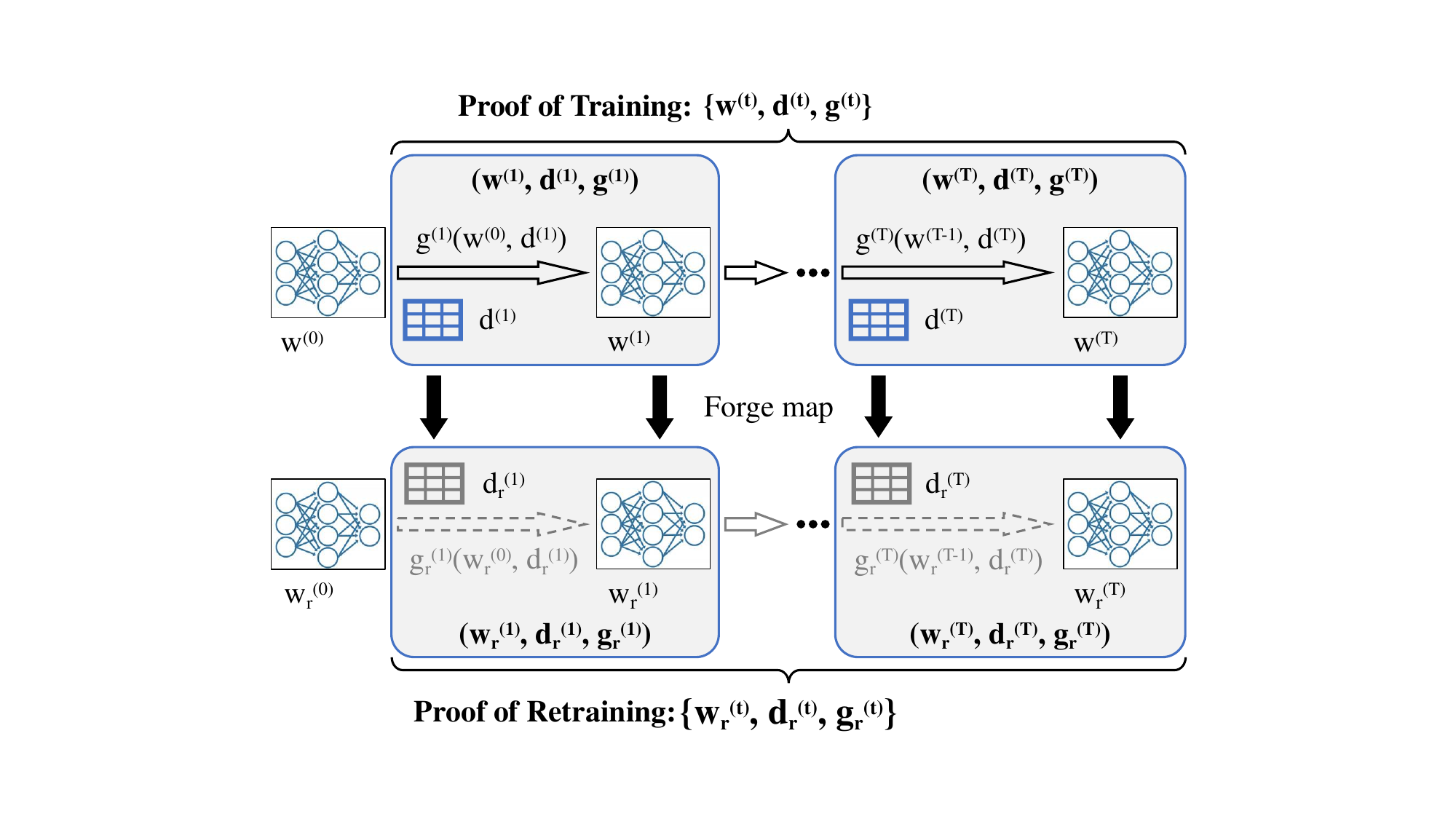}
    \vspace{-2em}
    \caption{An illustration of the forging-based adversarial unlearning framework. Different from the retraining-based adversarial method, the PoRT here is generated directly from the PoT recorded in the original training. $\bm{w}_r^{(t)}$ (with $d_r^{(t)}$) is obtained by conducting the forging map over the PoT instead of using the model updating function $g_r^{(t)}$.}
    \label{fig:forge}
    \vspace{-15pt}
\end{figure}

(1). For $t\in\mathcal{I}_e$, although the data $d^{(t)}$ requires no modification (we can simply let $d_r^{(t)}=d^{(t)}$), we still need to slightly alter the model parameters $\bm{w}^{(t)}$ because otherwise, the verifier might notice a large proportion of unchanged models comparing $\bm{w}_r^{(t)}$ with $\bm{w}^{(t)}$.
On the other hand, the majority of triplets should be updated carefully because the model utility should not decrease very much after multiple verification requests.
Considering both requirements for efficiency and model utility, we use single-step stochastic gradient descent (SGD) to update the model parameters in $\mathcal{F}^{(t)}$ and have
\begin{equation}\label{eq:w_update_e}
\bm{w}_r^{(t)}=\bm{w}^{(t)}-\gamma_r^{(t)}\nabla l(f_{\bm{w}^{(t)}}(\bm{x}^{(t)}),y^{(t)}), 
\end{equation}
where $(\bm{x}^{(t)},y^{(t)})\sim\mathcal{D}\backslash\mathcal{D}_u$ is a random sample chosen from the retained data and $\gamma_r^{(t)}$ is a small value to control the verification error $\|\bm{w}_r^{(t)}-g_r^{(t)}(\bm{w}_r^{(t-1)},d_r^{(t)})\|\leq\varepsilon$.

(2). For $t\in\mathcal{I}_n$, the forging map should remove the unlearned data from $d^{(t)}$.
To reduce the verification error, we still exploit the same strategy in our first adversarial method to select the mini-batch $d_r^{(t)}\in\mathcal{D}\backslash\mathcal{D}_u$ with $\mathcal{S}_n(d^{(t)})$ (we do not use $\mathcal{S}_r$ due to the efficiency issue).
Consequently, to update the model parameters in the forging map $\mathcal{F}^{(t)}$, we let
\begin{equation}\label{eq:w_update_n}
\bm{w}_r^{(t)}=g^{(t)}(\bm{w}^{(t-1)},\mathcal{S}_n(d^{(t)})).
\end{equation}

\begin{algorithm}[t]
   \caption{Forging-based Adversarial Unlearning Algorithm}
   \label{alg:adv2}
\begin{algorithmic}
   \STATE {\bfseries Input:} Training data $\mathcal{D}$, unlearned data $\mathcal{D}_u$, closest-neighbor mapping $\mathcal{N}:\mathcal{D}_u\rightarrow\mathcal{D}\backslash\mathcal{D}_u$, Proof of Training $\mathcal{P}_t=\{\bm{w}^{(t)},d^{(t)},g^{(t)}\}$.
   \STATE {\bfseries Output:} Proof of Retraining $\mathcal{P}_r$.
   \STATE $\mathcal{P}_r\leftarrow\emptyset$.
   \FOR{$t=1$ {\bfseries to} $T$ \textbf{in parallel}} 
   \IF{$d^{(t)}\cap\mathcal{D}_u=\emptyset$}
   \STATE $d_r^{(t)}\leftarrow d^{(t)}$.
   \STATE $\bm{w}_r^{(t)}\leftarrow\bm{w}^{(t)}-\gamma_r^{(t)}\nabla l(f_{\bm{w}^{(t)}}(\bm{x}^{(t)}),y^{(t)})$. 
   \ELSE
   \STATE $d_r^{(t)}\leftarrow\mathcal{S}_n(d^{(t)})$.
   \STATE $\bm{w}_r^{(t)}\leftarrow g^{(t)}(\bm{w}^{(t-1)},d_r^{(t)})$. 
   \ENDIF
   \STATE $g_r^{(t)}\leftarrow g^{(t)}$.
   \STATE $\mathcal{P}_r\leftarrow\mathcal{P}_r\cup(\bm{w}_r^{(t)},d_r^{(t)},g_r^{(t)})$.
   \ENDFOR
\end{algorithmic}
\end{algorithm}

We conclude our second adversarial unlearning process in \cref{alg:adv2} and provide a distinct illustration in \cref{fig:forge}.
The advantage of \cref{alg:adv2} is that the forging map can be implemented in parallel thanks to our formulation in \cref{eq:forging_map}, which largely reduces the execution time in real-world scenarios with frequent unlearning requests.
Specifically, we analyze the condition of \cref{alg:adv2} satisfying the reproducing verification and the time complexity of \cref{alg:adv2} as follows.
\begin{proposition}\label{pro:valid_2}
Under the same assumptions as in \cref{pro:cvg}, when $g_r^{(t)}$ is the vanilla SGD, 
$0\leq\gamma^{(t)}\leq\frac{1}{2L}(\sqrt{9+4\varepsilon L/(L_xC_D)}-3)$, and $0\leq\gamma_r^{(t)}\leq\frac{\varepsilon-\gamma^{(t)}L_xC_D}{G(2+\gamma^{(t)}L)}$,
\cref{alg:adv2} returns a valid Proof of Retraining under $\varepsilon>0$ threshold.
\end{proposition}
\begin{proposition}\label{pro:time_complexity}
Assume the time complexity of naive retraining is $T(n)$. The time complexity of \cref{alg:adv1} is $T(n)$ and the time complexity of \cref{alg:adv2} is $(\frac{p_u}{P}+\frac{1-p_u}{mP})\cdot T(n)$, where $p_u=1-(1-|\mathcal{D}_u|/|\mathcal{D}|)^m$, $P$ denotes the number of processes in parallel, and $m$ denotes the batch size.
\end{proposition}

The proofs of \cref{pro:valid_2} and \cref{pro:time_complexity} can be found in \cref{sec:proof}.
\cref{pro:valid_2} provides a theoretical guarantee when \cref{alg:adv2} successfully deceives the $\varepsilon$-threshold reproducing verification.
\cref{pro:time_complexity} compares the time complexity of \cref{alg:adv1} and \cref{alg:adv2} with naive retraining and demonstrates the superiority of \cref{alg:adv2} in computational efficiency.
Despite the efficiency of \cref{alg:adv2}, it can fail to satisfy the backdoor verification because the final unlearned model $\bm{w}_r^{(T)}$ is still dependent on the information of injected backdoor data.


\begin{table}[t]
    \centering
    \small
    \renewcommand{\arraystretch}{1.05}
    \caption{Dataset statistics.}\label{tab:dataset statistics}
    \aboverulesep = 0pt
    \belowrulesep = 0pt
    \begin{tabular}{l|cccc}
        Dataset & \# Train & \# Test & \# Class & Image Size \\
    \midrule
        MNIST & 60,000 & 10,000 & 10 & 28$\times$28 \\
        CIFAR-10 & 50,000 & 10,000 & 10 & 32$\times$32$\times$3 \\
        SVHN & 73,257 & 26,032 & 10 & 32$\times$32$\times$3 \\
    \end{tabular}
\vspace{-5mm}
\end{table}

\section{Experiments}
In this section, we empirically evaluate the vulnerability of MUL verification with numerical experiments. 
\paragraph{Datasets.}
Our experiments are based on three widely adopted real-world datasets for image classification, MNIST~\citep{lecun1998gradient}, SVHN~\citep{netzer2011reading}, and CIFAR-10~\citep{krizhevsky2009learning}: the MNIST dataset consists of a collection of handwritten digit images; the CIFAR-10 dataset contains color images in 10 classes, with each class representing a specific object category, e.g., cats and automobiles; the SVHN dataset consists of house numbers images captured from Google Street View.
The statistics of these three datasets are shown in \cref{tab:dataset statistics}.
All datasets are publicly accessible (MNIST with GNU General Public License, CIFAR-10 with MIT License, and SVHN with CC BY-NC License).
\paragraph{Evaluation Metrics.}
Recall that the goal of adversarial unlearning processes is to preserve the model utility and improve the efficiency of unlearning while circumventing the verification methods.
To demonstrate the efficacy of our adversarial unlearning methods, we use the verification error threshold $\varepsilon$ to evaluate the forging-based method and use the probability of type II errors (the target data is not unlearned is regarded as the null hypothesis) to evaluate our retraining-based method.
In addition, we should also ensure that adversarial unlearning methods benefit model providers in preserving model utility and improving efficiency.
Hence, we use utility metrics (e.g., F1-score) and the execution time to verify the benefit of adversarial unlearning methods.
\paragraph{Implementation.}
We implemented all experiments in the PyTorch~\citep{paszke2019pytorch} library and exploited SGD as the optimizer for training.
All experiments were conducted on an Nvidia RTX A6000 GPU.
We use the verification error threshold $\varepsilon$ to evaluate the forging-based method and use the probability of type II errors (the verifier thinks target data is unlearned but actually not) to evaluate the retraining-based method.
We use utility metrics (e.g., F1-score) and the execution time to show the benefit of adversarial unlearning.
We report the average value and standard deviation of the numerical results under five random seeds. 
More details of the hyperparameter setting are presented in \cref{sec:experiment}.
Our code is available at \url{https://github.com/zhangbinchi/unlearning-verification-is-fragile}.

\begin{figure}[t]
    \centering
    \subfigure[Verification error]{
    \includegraphics[width=0.48\linewidth]{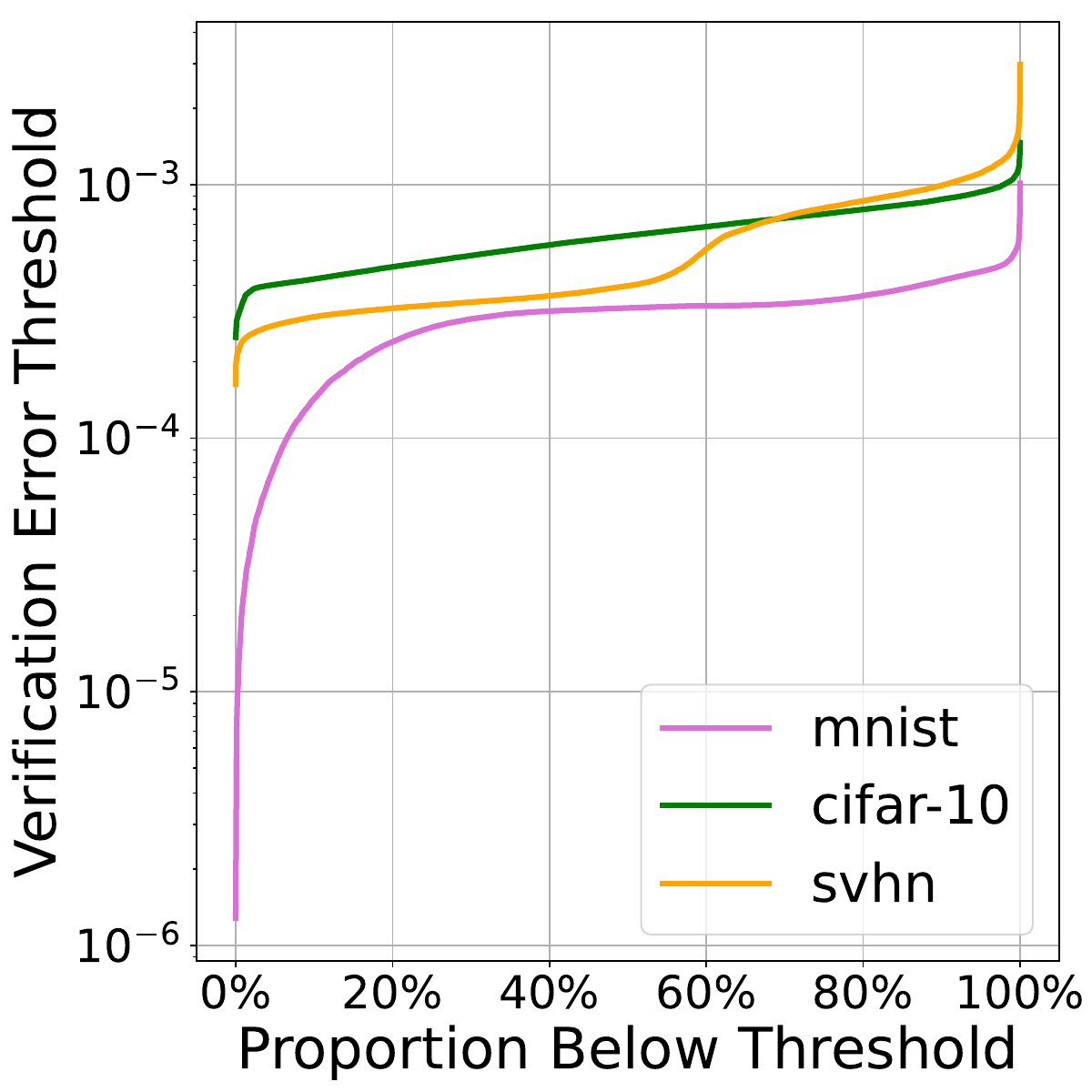}\label{fig:verif_error}}
    \subfigure[Details of error statistics]{
    \includegraphics[width=0.48\linewidth]{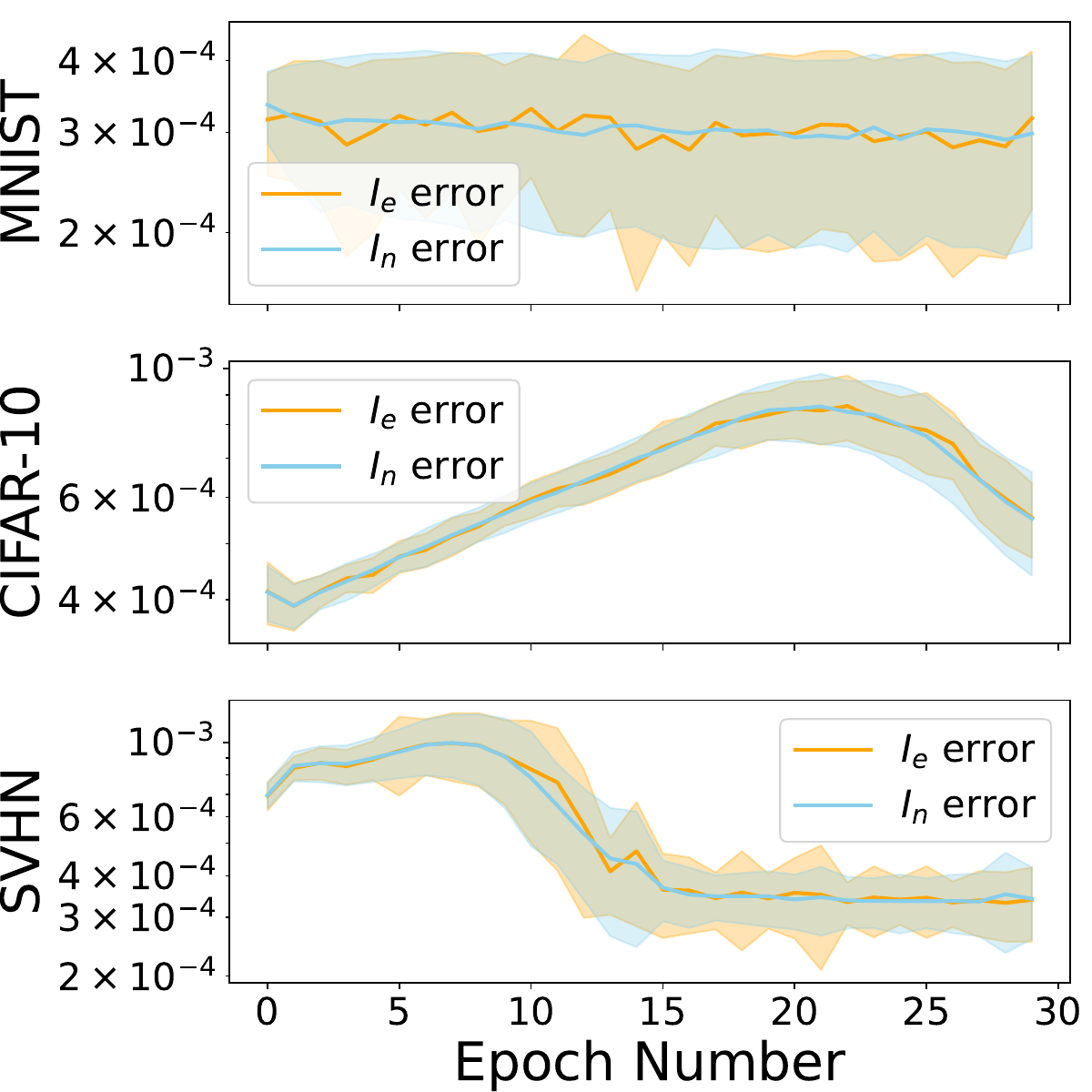}\label{fig:error_statistics}}
    \vspace{-5mm}
    \caption{Verification error of forging-based adversarial unlearning method for MLP over MNIST, CNN over CIFAR-10, and ResNet over SVHN.}
    \label{fig:verify_error}
    \vspace{-5mm}
\end{figure}

\begin{table}[t]
\small
\renewcommand{\arraystretch}{1.2}
\centering
\caption{Probability of the type II error ($\beta$ value) of backdoor verification on different (adversarial) unlearning strategies over three datasets.}
\label{tab:backdoor}
\vspace{1mm}
\aboverulesep = 0pt
\belowrulesep = 0pt
\begin{tabular}{c|ccc}
\toprule
\textbf{Method} & \textbf{MNIST} & \textbf{CIFAR-10} & \textbf{SVHN} \\
\midrule
Original & $2.61\times10^{-42}$ &$5.14\times10^{-20}$  & $1.11\times10^{-28}$ \\
Retrain &0.998  & 0.998 & 0.999 \\
Adv-R &0.997 & 0.997 & 0.999 \\
Adv-F & $5.46\times10^{-34}$ &$2.78\times10^{-16}$  & $2.08\times10^{-26}$ \\
\bottomrule
\end{tabular}
\vspace{-5mm}
\end{table}

\subsection{Adversarial Unlearning vs Verification}
We first empirically demonstrate the efficacy of our proposed adversarial unlearning methods, i.e., they can satisfy the verification and let the verifier believe that the data is unlearned normally.
We next discuss the reproducing verification and the backdoor verification separately.

\paragraph{Reproducing Verification.}
For reproducing verification, our retraining-based method can always satisfy the verification with an exact $0$ verification error because the PoRT is recorded directly based on the retraining process.
Hence, we mainly focus on the efficacy of the forging-based method in this experiment.
As the choice of verification error threshold $\varepsilon$ is highly personalized, we directly fix the hyperparameters of our forging-based method and record the verification error in each model updating step.
Specifically, we randomly choose 2\% data as the unlearned set. 
The hyperparameter settings are shown in \cref{sec:experiment}, and the experimental results are shown in \cref{fig:verify_error}.
We can observe that the verification errors in most steps for all datasets are below $1e^{-3}$, i.e., our forging-based method can satisfy the reproducing verification with a $1e^{-3}$ threshold.
Compared with the scale of adopted neural networks, $1e^{-3}$ can be seen as a small value.
We can also see a reduction in the overall verification error as the model scale decreases (from ResNet to MLP).
In addition, we also illustrate the mean value and the standard deviation of the verification error of $\mathcal{I}_e$ (excluding $\mathcal{D}_u$) and $\mathcal{I}_n$ (including $\mathcal{D}_u$) steps in different epochs and obtain the following observations: 1) the verification errors in $\mathcal{I}_e$ and $\mathcal{I}_n$ steps are consistent, perhaps due to the high dependence of both errors on the gradient scale; 2) within the overall threshold, the verification error tends to first increase and then decrease until staying nearby a local optimum; again, we attribute this to the dependence of errors on gradients.

\begin{table*}[t]
\small
\renewcommand{\arraystretch}{1.1}
\tabcolsep = 1.2pt
\centering
\caption{Comparison of the model utility among original training, naive retraining, and adversarial unlearning methods over three popular DNNs across three real-world datasets. We record the macro F1-score of the predictions on the unlearned set $\mathcal{D}_u$, retained set $\mathcal{D}\backslash\mathcal{D}_u$, and test set $\mathcal{D}_t$. The prefix `im-' denotes the results in the class-imbalanced setting.}
\label{tab:utility}
\aboverulesep = 0pt
\belowrulesep = 0pt
\begin{tabular}{c|ccc|ccc|ccc}
\toprule
\multirow{2}{*}{\textbf{Method}} & \multicolumn{3}{c|}{\textbf{MLP \& MNIST}} & \multicolumn{3}{c|}{\textbf{CNN \& CIFAR-10}} & \multicolumn{3}{c}{\textbf{ResNet \& SVHN}} \\
& $\mathcal{D}_u$ & $\mathcal{D}\backslash\mathcal{D}_u$ & $\mathcal{D}_t$ & $\mathcal{D}_u$ & $\mathcal{D}\backslash\mathcal{D}_u$ & $\mathcal{D}_t$ & $\mathcal{D}_u$ & $\mathcal{D}\backslash\mathcal{D}_u$ & $\mathcal{D}_t$ \\
\midrule
Original & 99.47 {\scriptsize $\pm$ 0.09} & 99.76 {\scriptsize $\pm$ 0.08} & 97.00 {\scriptsize $\pm$ 0.17} & 100.00 {\scriptsize $\pm$ 0.00} & 100.00 {\scriptsize $\pm$ 0.00} & 85.33 {\scriptsize $\pm$ 0.31} & 100.00 {\scriptsize $\pm$ 0.00} & 100.00 {\scriptsize $\pm$ 0.00} & 94.91 {\scriptsize $\pm$ 0.09} \\
Retrain & 96.43 {\scriptsize$\pm$ 0.19} & 99.52 {\scriptsize$\pm$ 0.10} & 96.75 {\scriptsize$\pm$ 0.13} & 83.60 {\scriptsize$\pm$ 0.31} & 100.00 {\scriptsize$\pm$ 0.00} & 83.12 {\scriptsize$\pm$ 0.23} & 94.33 {\scriptsize$\pm$ 0.24} & 100.00 {\scriptsize$\pm$ 0.00} & 94.57 {\scriptsize$\pm$ 0.06} \\
Adv-R ($\mathcal{S}_r$) & 98.17 {\scriptsize$\pm$ 0.16} & 99.33 {\scriptsize$\pm$ 0.18} & 96.78 {\scriptsize$\pm$ 0.13} & 83.81 {\scriptsize$\pm$ 0.44} & 100.00 {\scriptsize$\pm$ 0.00} & 83.08 {\scriptsize$\pm$ 0.34} & 94.38 {\scriptsize$\pm$ 0.11} & 100.00 {\scriptsize$\pm$ 0.00} & 94.54 {\scriptsize$\pm$ 0.09} \\
Adv-R ($\mathcal{S}_n$) & 96.34 {\scriptsize$\pm$ 0.11} & 98.65 {\scriptsize$\pm$ 0.19} & 96.60 {\scriptsize$\pm$ 0.14} & 82.40 {\scriptsize$\pm$ 0.39} & 100.00 {\scriptsize$\pm$ 0.00} & 81.85 {\scriptsize$\pm$ 0.44} & 94.64 {\scriptsize$\pm$ 0.20} & 100.00 {\scriptsize$\pm$ 0.00} & 94.75 {\scriptsize$\pm$ 0.04} \\
Adv-F & 99.30 {\scriptsize$\pm$ 0.13} & 99.33 {\scriptsize$\pm$ 0.10} & 96.94 {\scriptsize$\pm$ 0.14} & 100.00 {\scriptsize$\pm$ 0.00} & 100.00 {\scriptsize$\pm$ 0.00} & 85.20 {\scriptsize$\pm$ 0.24} & 100.00 {\scriptsize $\pm$ 0.00} & 100.00 {\scriptsize $\pm$ 0.00} & 94.91 {\scriptsize $\pm$ 0.07} \\
\midrule
im-Original & 60.29 {\scriptsize $\pm$ 13.07} & 97.00 {\scriptsize $\pm$ 2.60} & 96.88 {\scriptsize $\pm$ 0.07} & 100.00 {\scriptsize $\pm$ 0.00} & 100.00 {\scriptsize $\pm$ 0.00} & 85.44 {\scriptsize $\pm$ 0.22} & 100.00 {\scriptsize $\pm$ 0.00} & 100.00 {\scriptsize $\pm$ 0.00} & 94.66 {\scriptsize $\pm$ 0.30} \\
im-Retrain & 38.76 {\scriptsize$\pm$ 13.41} & 95.86 {\scriptsize$\pm$ 4.34} & 89.92 {\scriptsize$\pm$ 5.70} & 24.25 {\scriptsize$\pm$ 6.98} & 90.88 {\scriptsize$\pm$ 5.03} & 65.22 {\scriptsize$\pm$ 5.94} & 33.08 {\scriptsize$\pm$ 11.97} & 95.19 {\scriptsize$\pm$ 3.78} & 83.89 {\scriptsize$\pm$ 5.83} \\
im-Adv-R ($\mathcal{S}_r$) & 39.48 {\scriptsize$\pm$ 12.20} & 99.71 {\scriptsize$\pm$ 0.19} & 91.04 {\scriptsize$\pm$ 4.80} & 25.94 {\scriptsize$\pm$ 6.89} & 96.00 {\scriptsize$\pm$ 4.90} & 76.51 {\scriptsize$\pm$ 4.32} & 34.76 {\scriptsize$\pm$ 12.06} & 98.00 {\scriptsize$\pm$ 4.01} & 87.63 {\scriptsize$\pm$ 6.11} \\
im-Adv-R ($\mathcal{S}_n$) & 42.90 {\scriptsize$\pm$ 11.87} & 97.96 {\scriptsize$\pm$ 0.59} & 92.80 {\scriptsize$\pm$ 4.54} & 23.97 {\scriptsize$\pm$ 8.75} & 91.46 {\scriptsize$\pm$ 1.81} & 67.34 {\scriptsize$\pm$ 4.02} & 33.92 {\scriptsize$\pm$ 11.85} & 99.37 {\scriptsize$\pm$ 0.20} & 84.87 {\scriptsize$\pm$ 5.42} \\
im-Adv-F & 64.21 {\scriptsize$\pm$ 9.89} & 97.28 {\scriptsize$\pm$ 3.34} & 96.81 {\scriptsize$\pm$ 0.04} & 100.00 {\scriptsize$\pm$ 0.00} & 100.00 {\scriptsize$\pm$ 0.00} & 85.11 {\scriptsize$\pm$ 0.21} & 100.00 {\scriptsize $\pm$ 0.00} & 100.00 {\scriptsize $\pm$ 0.00} & 94.77 {\scriptsize $\pm$ 0.09} \\
\bottomrule
\end{tabular}
\vspace{-10pt}
\end{table*}

\paragraph{Backdoor Verification.}
For backdoor verification, we exploit Athena~\citep{sommer2022athena} as the verification strategy. 
In particular, we randomly choose the backdoor training and test data, inject a specific pattern of pixels, and change the label of training data to a fixed target label.
To test the integrity of unlearning, we make unlearned data include the backdoor training data.
If the backdoor success rate is close to the random prediction, the model can predict the original labels for the backdoor test data, and the unlearned model was not trained on the poisoned data. Thus, the verifier believes that the model provider follows the unlearning request. 
Specifically, we regard ``the target data is unlearned" as the null hypothesis and ``the target data is not unlearned" as the alternative hypothesis. 
We use the probability of type II errors (the verifier thinks target data is unlearned but actually not) to evaluate our adversarial unlearning methods.
The results are shown in \cref{tab:backdoor}.
Consistent with our former discussion, the retraining-based adversarial method and naive retraining are almost always regarded as truly unlearning the target data; the forging-based adversarial method and original training are hardly regarded as truly unlearning the target data.

\begin{figure}[t]
    \centering
    \includegraphics[width=0.8\linewidth]{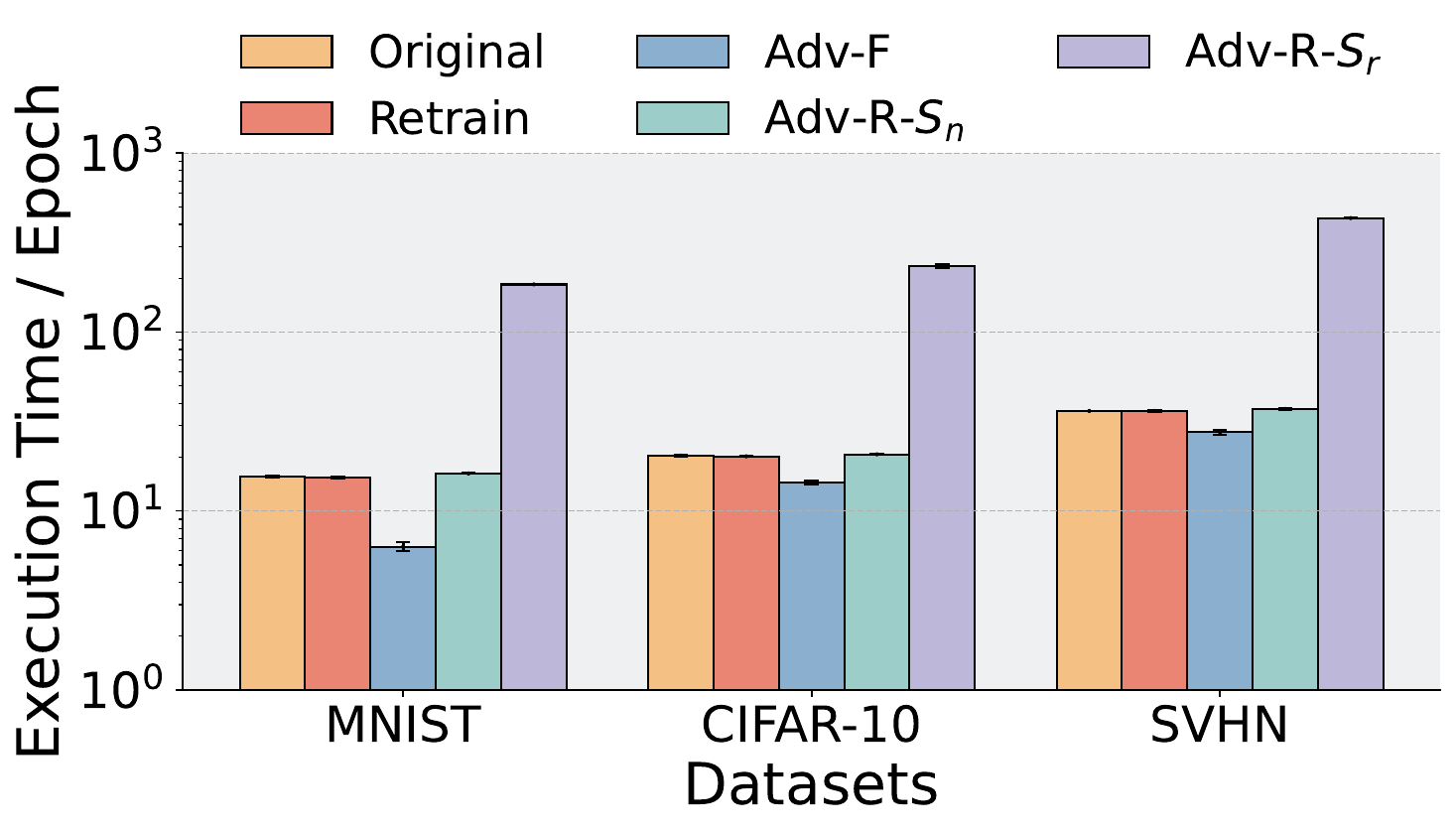}
    \vspace{-8pt}
    \caption{Comparison of execution time among original training, naive retraining, and adversarial unlearning methods over three real-world datasets.}
    \label{fig:time}
    \vspace{-10pt}
\end{figure}

\subsection{Adversary's Goal}\label{sec:Adv goal}
In previous experiments, we have demonstrated that our proposed adversarial unlearning methods can satisfy reproducing and backdoor verification.
Next, we aim to show that 1) the adversarial unlearning deceives the verification: they still memorize the information of unlearned data, and 2) model providers benefit from adversarial unlearning: they preserve the model utility and improve unlearning efficiency.

\paragraph{Utility.}
In this experiment, we compare the model utility between naive retraining and our adversarial unlearning. 
To simulate the data heterogeneity in real-world scenarios, we add a class-imbalanced unlearning setting. 
For the retraining-based adversarial method, we adopt both random sampling $\mathcal{S}_r$ and nearest neighbor $\mathcal{S}_n$ approaches to select mini-batches.
For the forging-based adversarial method, we use the last recorded model on the PoRT $\bm{w}_r^{(T)}$ for utility evaluation.
To demonstrate the long-term effect of adversarial unlearning, we divide 10\% of the training data as the unlearned set.
Details of hyperparameter settings and complete experimental results can be found in \cref{sec:utl_exp}, and we provide the truncated experimental results in \cref{tab:utility}.
From the results, we can obtain the following observations: 
1) in the normal setting, the retraining-based adversarial method has a similar utility to naive retraining;
2) in the class-imbalanced setting, the retraining-based adversarial method has a much better utility than naive retraining, which means model providers can benefit more when data heterogeneity exists in the unlearning process.
3) in both settings, the forging-based adversarial method has the best utility of the unlearned model (close to the original training).

\begin{figure*}[t]
    \centering
    \subfigure[Selection time of two mini-batch selection strategies.]{
    \includegraphics[width=0.3\linewidth]{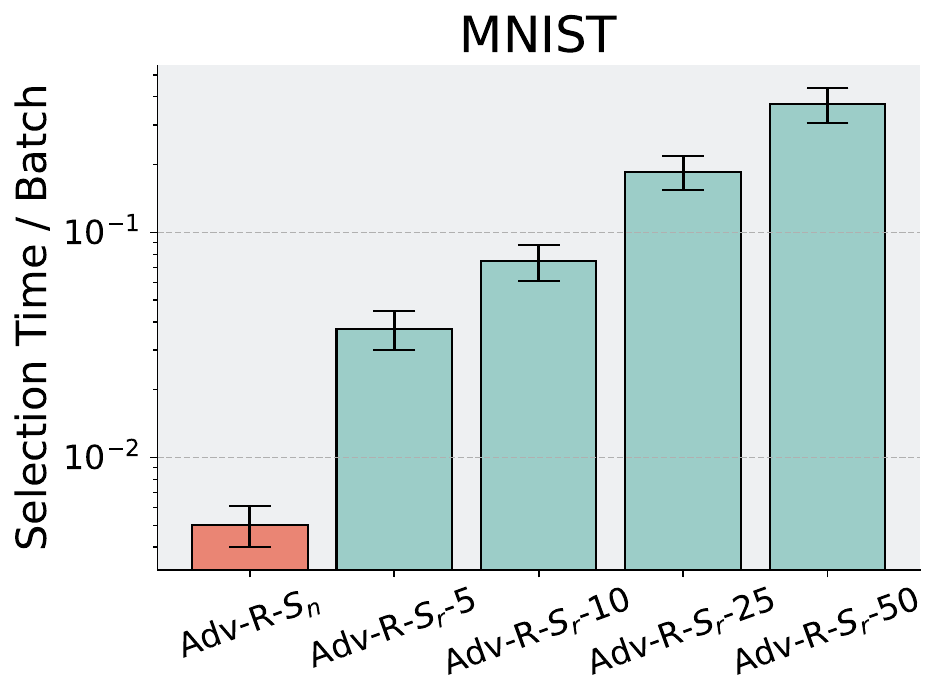}\label{fig:mnist_time_perbatch}}
    \subfigure[Gradient distance between mini-batch selection strategies and the original batch.]{
    \includegraphics[width=0.3\linewidth]{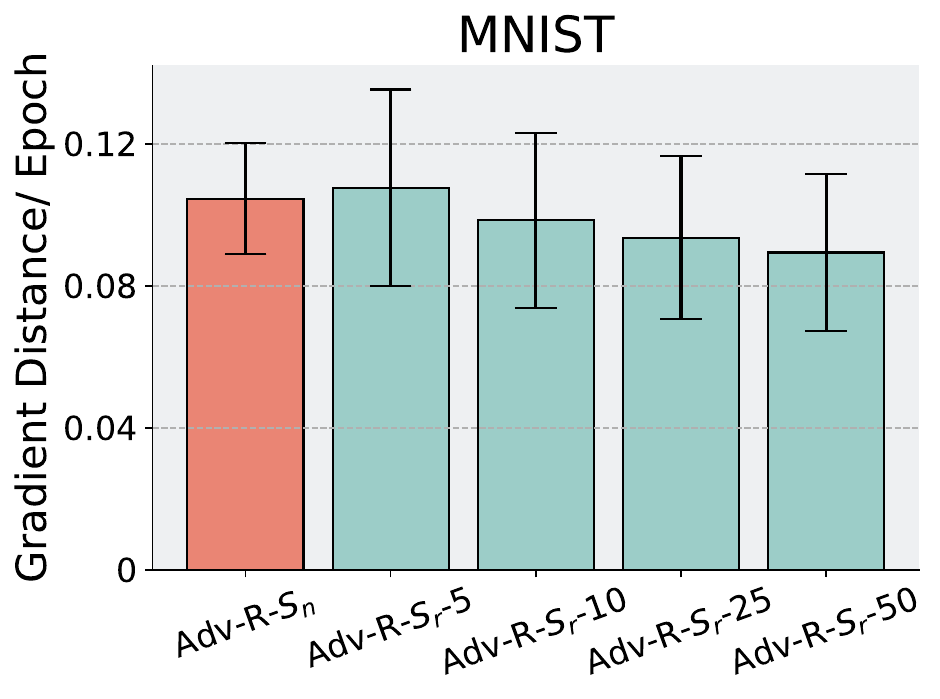}\label{fig:mnist_error_perepoch}}
    \subfigure[Gradient distance in different datasets.]{
    \includegraphics[width=0.3\linewidth]{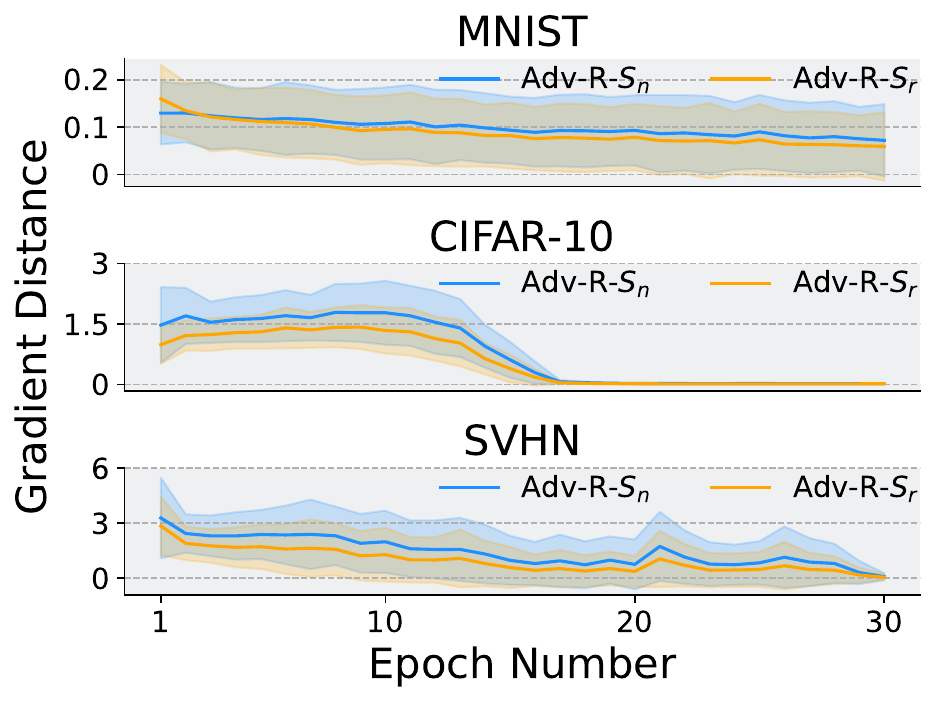}\label{fig:random_nearest_error}}
    \vspace{-3mm}
    \caption{Comparison of two mini-batch selection strategies: random sampling $\mathcal{S}_r$ and nearest neighbor $\mathcal{S}_n$.}
    \label{fig:batch_selection}
\end{figure*}

\paragraph{Efficiency.}
In this experiment, we compare the computational efficiency between naive retraining and our adversarial unlearning.
For the retraining-based adversarial method, we use the nearest neighbor selection to choose the mini-batches and obtain the nearest neighbor mapping by ranking the pre-computed distance between the unlearned data and the retained data.
For the forging-based adversarial method, we adopt five processes in parallel to compute the PoRT.
We record the average execution time for each epoch and show the results in \cref{fig:time}.
From the results, we can obtain that 1) the forging-based method has the shortest execution time; 2) the time cost of forging in practice can be longer than theoretical results due to the resource bottleneck of the adopted single GPU; 3) the retraining-based method is slightly slower than naive retraining due to the nearest neighbor calculation, but our proposed nearest neighbor selection is much faster than random sampling.

\subsection{Mini-batch Selection: Nearest Neighbor vs Random Sampling}
In~\cref{sec:retrain based}, we introduced two mini-batch selection strategies for the retraining-based method: random sampling $\mathcal{S}_r$ and nearest neighbor $\mathcal{S}_n$. Additionally, in~\cref{sec:Adv goal}, we demonstrated that our nearest neighbor selection strategy is faster than random sampling. This section delves deeper into comparing these two strategies.

\cref{fig:mnist_error_perepoch} and~\cref{fig:mnist_time_perbatch} illustrate the selection time per epoch and the gradient distance between selected batches and their corresponding batches containing unlearn data for both strategies. We conduct these experiments on the MNIST dataset, varying the number of batch samples $M$ for random sampling. In this strategy, $M$ candidate batches are selected, and the one with the smallest gradient distance to the original batch is chosen. We tested $M$ values of 5, 10, 25, and 50.

From~\cref{fig:mnist_time_perbatch}, we observe that: i) in the random sampling strategy, selection time increases with $M$ due to the increased gradient computations; ii) our nearest neighbor selection consistently outperforms random sampling in speed, even with smaller $M$ values, by avoiding extra gradient computations. \cref{fig:mnist_error_perepoch} shows that: i) larger $M$ values in random sampling lead to better batch selection; ii) nearest neighbor selection performs better than random sampling with small $M$ values and is comparable when $M$ is large. Further, we extended our experiments to two additional datasets with $M=50$ for the random sampling strategy. As depicted in~\cref{fig:random_nearest_error}, our strategy exhibits similar performance to random sampling. In summary, the nearest neighbor strategy offers significantly faster selection times and comparable model training performance compared to the random sampling strategy.

\section{Conclusion}
In this paper, we expose the vulnerability in the verification of MUL.
In particular, we summarize current verification strategies into two types.
Regarding both types, we propose two adversarial unlearning processes that circumvent the verification while preserving the information of unlearned data.
Our theoretical and empirical results highlight the following conclusions:
the adversarial unlearning method can circumvent the verification methods while improving the model utility without extra computation costs. 
Put another way of thinking, by adopting an adversarial unlearning method, a dishonest model provider gains a higher model utility (still memorizes the unlearned data) after unlearning without compromising efficiency. 
Furthermore, the retraining-based adversarial method can perfectly circumvent all existing verification methods relying on the natural similarity within the training data and the randomness of the training algorithm. 
This threat poses a novel challenge for the safe verification of machine unlearning: there is no strict guarantee for the verification of MUL.
Better verification methods are needed to obtain precise and reliable verification results.

\section*{Acknowledgements}

This work is supported in part by the National Science Foundation under grants (IIS-2006844, IIS-2144209, IIS-2223769, CNS-2154962, BCS-2228534, ECCS-2033671, ECCS-2143559, CPS-2313110, and SII-2132700), the Commonwealth Cyber Initiative Awards under grants (VV-1Q23-007, HV-2Q23-003, and VV-1Q24-011), the JP Morgan Chase Faculty Research Award, and the Cisco Faculty Research Award.

\section*{Impact Statement}
Our proposed adversarial unlearning methods in this paper can threaten the safety of the verification of MUL.
Generally, we can consider that a model provider trains an ML model based on some personal data collected from data owners.
According to the legislation~\citep{CCPA,GDPR}, the data owners have the right to have their data removed from the ML model and take actions to verify the efficacy of unlearning.
However, using our proposed adversarial unlearning algorithms, the model provider can make the data owners (verifiers) believe that their data has been unlearned while preserving the information of these personal data.

Despite that, our research has a more significant positive influence compared with the potential risks. 
The study of MUL verification is still at a nascent stage.
Given the deficient understanding of the safety of MUL verification, our study highlights the vulnerability of current verification strategies of MUL and inspires further research on the safe verification of MUL.
Moreover, we tentatively provide discussions on the weak points of our proposed adversarial unlearning strategies and insights for detecting these misbehaviors in \cref{sec:defense}.


\bibliography{icml2024}
\bibliographystyle{icml2024}

\newpage
\appendix
\onecolumn
\section{Proof}\label{sec:proof}

\subsection{Proof of \cref{pro:valid}}
\begin{proof}
To prove that \cref{alg:adv1} returns a valid PoRT, we verify the two properties of a valid PoRT in \cref{def:valid_proof}.
First, we verify the removable property.
For any $d\in\mathcal{S}_r(\bm{w}_r^{(t-1)};d^{(t)})\cup\mathcal{S}_n(d^{(t)})$, we have $d\in\mathcal{D}\backslash\mathcal{D}_u$. Hence, we have $d_r^{(t)}\cap\mathcal{D}_u=\emptyset$.
We then verify the reproducible property.
It is obvious that $\|\bm{w}_r^{(t)}-g_r^{(t)}(\bm{w}_r^{(t-1)},d_r^{(t)})\|=0$ and the reproducible property holds for $\varepsilon=0$.
In conclusion, \cref{alg:adv1} returns a valid PoRT.
\end{proof}

\subsection{Proof of \cref{pro:cvg}}\label{sec:converge}
\begin{proof}
Recall that \cref{alg:adv1} can be seen as a mini-batch SGD with the following model updating function:
\begin{equation}
\begin{aligned}
\bm{w}_r^{(t+1)}&=\bm{w}_r^{(t)}-\gamma\nabla\mathcal{L}(\bm{w}_r^{(t)},d_r^{(t+1)}) \\
&=\bm{w}_r^{(t)}-\gamma\nabla\mathcal{L}(\bm{w}_r^{(t)},\mathcal{S}_n(d^{(t+1)})),
\end{aligned}
\end{equation}
where $d^{(t+1)}\sim\mathcal{D}$.
It is worth noting that $d^{(t+1)}$ is randomly chosen from $\mathcal{D}$, while $d_r^{(t+1)}$ relies on the selection of $d^{(t+1)}$ and the unlearned set $\mathcal{D}_u$.
If we take the expectation value over both sides, it will be difficult to directly compute the expectation $\mathrm{E}[\nabla\mathcal{L}(\bm{w}_r^{(t)},d_r^{(t+1)})]$ on the right-hand side.
Consequently, we let $\bm{b}^{(t)}=\nabla\mathcal{L}(\bm{w}_r^{(t)},d_r^{(t+1)})-\nabla\mathcal{L}(\bm{w}_r^{(t)},d^{(t+1)})$ and have
\begin{equation}
\bm{w}_r^{(t+1)}=\bm{w}_r^{(t)}-\gamma(\nabla\mathcal{L}(\bm{w}_r^{(t)},d^{(t+1)})+\bm{b}^{(t)}).
\end{equation}
For simplicity, we denote $\mathcal{L}(\bm{w})$ as the loss over the whole training set $\mathcal{L}(\bm{w},\mathcal{D})$ and $\mathcal{L}_t(\bm{w})$ as the loss over the original mini-batch at the $t$-th iteration $\mathcal{L}(\bm{w},d^{(t+1)})$.
Next, we focus on the loss function value
\begin{equation}
\begin{aligned}
\mathcal{L}(\bm{w}_r^{(t+1)})&=\mathcal{L}(\bm{w}_r^{(t)}-\gamma(\nabla\mathcal{L}_t(\bm{w}_r^{(t)})+\bm{b}^{(t)})) \\
&=\mathcal{L}(\bm{w}_r^{(t)})-\nabla\mathcal{L}(\bm{w}_r^{(t)})^\top\gamma(\nabla\mathcal{L}_t(\bm{w}_r^{(t)})+\bm{b}^{(t)})+\frac{\gamma^2}{2}(\nabla\mathcal{L}_t(\bm{w}_r^{(t)})+\bm{b}^{(t)})^\top\nabla^2\mathcal{L}(\bm{\xi}^{(t)})(\nabla\mathcal{L}_t(\bm{w}_r^{(t)})+\bm{b}^{(t)}) \\
&\leq\mathcal{L}(\bm{w}_r^{(t)})-\gamma\nabla\mathcal{L}(\bm{w}_r^{(t)})^\top(\nabla\mathcal{L}_t(\bm{w}_r^{(t)})+\bm{b}^{(t)})+\frac{\gamma^2L}{2}\|\nabla\mathcal{L}(\bm{w}_r^{(t)})+\bm{b}^{(t)}\|^2 \\
\end{aligned}
\end{equation}
The second equality sign is based on the first-order Taylor's theorem with $\frac{\gamma^2}{2}(\nabla\mathcal{L}_t(\bm{w}_r^{(t)})+\bm{b}^{(t)})^\top\nabla^2\mathcal{L}(\bm{\xi}^{(t)})(\nabla\mathcal{L}_t(\bm{w}_r^{(t)})+\bm{b}^{(t)})$ as the Lagrange remainder.
Given $\bm{w}_r^{(t)}$, we take the expectation over the randomly selected mini-batch $d^{(t)}$ for both sides and obtain
\begin{equation}\label{eq:expectation inequality}
\begin{aligned}
\mathrm{E}[\mathcal{L}(\bm{w}_r^{(t+1)})]&\leq\mathrm{E}[\mathcal{L}(\bm{w}_r^{(t)})-\gamma\nabla\mathcal{L}(\bm{w}_r^{(t)})^\top(\nabla\mathcal{L}_t(\bm{w}_r^{(t)})+\bm{b}^{(t)})+\frac{\gamma^2L}{2}\|\nabla\mathcal{L}(\bm{w}_r^{(t)})+\bm{b}^{(t)}\|^2] \\
&\leq\mathrm{E}[\mathcal{L}(\bm{w}_r^{(t)})]-\gamma\mathrm{E}[\nabla\mathcal{L}(\bm{w}_r^{(t)})^\top\nabla\mathcal{L}_t(\bm{w}_r^{(t)})]-\gamma\mathrm{E}[\nabla\mathcal{L}(\bm{w}_r^{(t)})^\top\bm{b}^{(t)}]+\frac{\gamma^2L}{2}\mathrm{E}[\|\nabla\mathcal{L}(\bm{w}_r^{(t)})+\bm{b}^{(t)}\|^2]) \\
&\leq\mathrm{E}[\mathcal{L}(\bm{w}_r^{(t)})]-\gamma\|\nabla\mathcal{L}(\bm{w}_r^{(t)})\|^2-\gamma(1-\gamma L)\mathrm{E}[\nabla\mathcal{L}(\bm{w}_r^{(t)})^\top\bm{b}^{(t)}]+\frac{\gamma^2L}{2}(G^2+\mathrm{E}[\|\bm{b}^{(t)}\|^2]).
\end{aligned}
\end{equation}
For each iteration $t$, we have one corresponding inequality as \cref{eq:expectation inequality}. We then sum up the \cref{eq:expectation inequality} for $t=0$ to $T-1$ and have
\begin{equation}\label{eq:inequality sum}
\sum_{t=0}^{T-1}\mathrm{E}[\mathcal{L}(\bm{w}_r^{(t+1)})]\leq\sum_{t=0}^{T-1}\mathrm{E}[\mathcal{L}(\bm{w}_r^{(t)})]-\gamma\|\nabla\mathcal{L}(\bm{w}_r^{(t)})\|^2-\gamma(1-\gamma L)\mathrm{E}[\nabla\mathcal{L}(\bm{w}_r^{(t)})^\top\bm{b}^{(t)}]+\frac{\gamma^2L}{2}(G^2+\mathrm{E}[\|\bm{b}^{(t)}\|^2]).
\end{equation}
After telescoping \cref{eq:inequality sum}, we have
\begin{equation}\label{eq:sum simplify}
\gamma\sum_{t=0}^{T-1}\|\nabla\mathcal{L}(\bm{w}_r^{(t)})\|^2+(1-\gamma L)\mathrm{E}[\nabla\mathcal{L}(\bm{w}_r^{(t)})^\top\bm{b}^{(t)}]\leq\mathcal{L}(\bm{w}_r^{(0)})-\mathcal{L}(\bm{w}_r^{(T)})+\frac{\gamma^2L}{2}\sum_{t=0}^{T-1}(G^2+\mathrm{E}[\|\bm{b}^{(t)}\|^2]).
\end{equation}
Without loss of generality, we specify the norm as $\ell$-2 norm and have 
\begin{equation}\label{eq:gradient sum}
\begin{aligned}
\|\bm{b}^{(t)}\|&=\|\nabla\mathcal{L}(\bm{w}_r^{(t)},d_r^{(t+1)})-\nabla\mathcal{L}(\bm{w}_r^{(t)},d^{(t+1)})\| \\
&=\frac{1}{m}\norm{\sum_{i=1}^m\nabla_{\bm{w}}l(\bm{w}_r^{(t)},d_{r_i}^{(t+1)})-\nabla_{\bm{w}}l(\bm{w}_r^{(t)},d_i^{(t+1)})} \\
&\leq\frac{1}{m}\sum_{i=1}^m\|\nabla_{\bm{w}}l(\bm{w}_r^{(t)},d_{r_i}^{(t+1)})-\nabla_{\bm{w}}l(\bm{w}_r^{(t)},d_i^{(t+1)})\|.
\end{aligned}
\end{equation}
In particular, we use $d_{r_i}^{(t+1)}$ and $d_i^{(t+1)}$ to denote the $i$-th data point in the mini-batch $d_{r}^{(t+1)}$ and $d^{(t+1)}$. 
We let $d_{r_i}^{(t+1)}=(\bm{x}_{r_i}^{(t+1)},y_{r_i}^{(t+1)})$ and $d_i^{(t+1)}=(\bm{x}_i^{(t+1)},y_i^{(t+1)})$, denoting the pair of feature vector and label.
It is worth noting that $y_{r_i}^{(t+1)}=y_i^{(t+1)}$ according to the definition of $\mathcal{S}_n$.
We denote $\bm{w}[j]$ as the $j$-th element of $\bm{w}$.
Given $y_{r_i}^{(t+1)}=y_i^{(t+1)}$, we omit the variable $y$ in $l(\cdot)$ and have 
\begin{equation}\label{eq:partial gradient}
\begin{aligned}
\left|\frac{\partial l(\bm{w}_r^{(t)},\bm{x}_{r_i}^{(t+1)})}{\partial\bm{w}[j]}-\frac{\partial l(\bm{w}_r^{(t)},\bm{x}_i^{(t+1)})}{\partial\bm{w}[j]}\right|&=\left|\int_{s=0}^1(\bm{x}_{r_i}^{(t+1)}-\bm{x}_i^{(t+1)})^\top\frac{\partial^2l(\bm{w}_r^{(t)},\bm{x}_i^{(t+1)}+s(\bm{x}_{r_i}^{(t+1)}-\bm{x}_i^{(t+1)}))}{\partial\bm{w}[j]\partial\bm{x}}ds\right| \\
&\leq\int_{s=0}^1\left|(\bm{x}_{r_i}^{(t+1)}-\bm{x}_i^{(t+1)})^\top\frac{\partial^2l(\bm{w}_r^{(t)},\bm{x}_i^{(t+1)}+s(\bm{x}_{r_i}^{(t+1)}-\bm{x}_i^{(t+1)}))}{\partial\bm{w}[j]\partial\bm{x}}\right|ds \\
&\leq\int_{s=0}^1\|\bm{x}_{r_i}^{(t+1)}-\bm{x}_i^{(t+1)}\|\cdot\norm{\frac{\partial^2l(\bm{w}_r^{(t)},\bm{x}_i^{(t+1)}+s(\bm{x}_{r_i}^{(t+1)}-\bm{x}_i^{(t+1)}))}{\partial\bm{w}[j]\partial\bm{x}}}ds \\
&\leq\int_{s=0}^1\|\bm{x}_{r_i}^{(t+1)}-\bm{x}_i^{(t+1)}\|\cdot\norm{\nabla_{\bm{x}}\frac{\partial l(\bm{w},\bm{x})}{\partial\bm{w}[j]}}ds \\
&=\|\bm{x}_{r_i}^{(t+1)}-\bm{x}_i^{(t+1)}\|\cdot\norm{\nabla_{\bm{x}}\frac{\partial l(\bm{w},\bm{x})}{\partial\bm{w}[j]}}.
\end{aligned}
\end{equation}
We then incorporate \cref{eq:partial gradient} into \cref{eq:gradient sum} and obtain that
\begin{equation}\label{eq:bounded bias}
\begin{aligned}
\|\bm{b}^{(t)}\|&\leq\frac{1}{m}\sum_{i=1}^m\|\nabla_{\bm{w}}l(\bm{w}_r^{(t)},d_{r_i}^{(t+1)})-\nabla_{\bm{w}}l(\bm{w}_r^{(t)},d_i^{(t+1)})\| \\
&\leq\frac{1}{m}\sum_{i=1}^mL_x\|\bm{x}_{r_i}^{(t+1)}-\bm{x}_i^{(t+1)}\| \\
&\leq L_xC_D=B.
\end{aligned}
\end{equation}
In addition, the unlearned data cannot be involved in every single iteration when $|\mathcal{D}_u|<T$.
When $d^{(t+1)}\cap\mathcal{D}_u=\emptyset$, we have $d^{(t+1)}=d_r^{(t+1)}$ and $\bm{b}^{(t)}=0$.
Considering that $d^{(t+1)}$ is chosen uniformly from $\mathcal{D}$, we can compute the probability of $d^{(t+1)}\cap\mathcal{D}_u\neq\emptyset$ as $p_u=1-(1-|\mathcal{D}_u|/|\mathcal{D}|)^m$.
Consequently, when $T\rightarrow\infty$, we can consider that only $p_uT$ iterations contain the bias term $\bm{b}^{(t)}$ and we have $\bm{b}^{(t)}=0$ for the remaining $(1-p_u)T$ iterations.
We then incorporate \cref{eq:bounded bias} back into \cref{eq:sum simplify} and have
\begin{equation}
\begin{aligned}
\gamma\sum_{t=0}^{T-1}\|\nabla\mathcal{L}(\bm{w}_r^{(t)})\|^2-\gamma(1-\gamma L)p_uTGB&\leq\gamma\sum_{t=0}^{T-1}\|\nabla\mathcal{L}(\bm{w}_r^{(t)})\|^2+(1-\gamma L)\mathrm{E}[\nabla\mathcal{L}(\bm{w}_r^{(t)})^\top\bm{b}^{(t)}] \\
&\leq\mathcal{L}(\bm{w}_r^{(0)})-\mathcal{L}(\bm{w}_r^{(T)})+\frac{\gamma^2L}{2}\sum_{t=0}^{T-1}(G^2+\mathrm{E}[\|\bm{b}^{(t)}\|^2]) \\
&\leq\mathcal{L}(\bm{w}_r^{(0)})-\mathcal{L}(\bm{w}_r^{(T)})+\frac{\gamma^2LT}{2}(G^2+p_uB^2).
\end{aligned}
\end{equation}
Finally, we can finish the proof by
\begin{equation}
\begin{aligned}
\mathrm{E}_T[\|\nabla\mathcal{L}(\bm{w}_r^{(T)})\|^2]&=\frac{1}{T}\sum_{t=0}^{T-1}\|\nabla\mathcal{L}(\bm{w}_r^{(t)})\|^2 \\
&\leq\frac{\mathcal{L}(\bm{w}_r^{(0)})-\mathcal{L}(\bm{w}_r^{(T)})}{\gamma T}+\frac{\gamma L}{2}(G^2+p_uB^2)+(1-\gamma L)p_uGB.
\end{aligned}
\end{equation}
\end{proof}

\subsection{Proof of \cref{pro:valid_2}}
\begin{proof}
To prove that \cref{alg:adv2} returns a valid PoRT, we verify the two properties of a valid PoRT in \cref{def:valid_proof}.
First, we verify the removable property.
For the $t$-th iteration, if $t\in\mathcal{I}_e$, we have that $d_r^{(t)}\cap\mathcal{D}_u=d^{(t)}\cap\mathcal{D}_u=\emptyset$ holds; if $t\in\mathcal{I}_e$, we have that 
\begin{equation}
\begin{aligned}
d_r^{(t)}\cap\mathcal{D}_u&=\mathcal{S}_n(d^{(t)})\cap\mathcal{D}_u \\
&=\left((d^{(t)}\backslash\mathcal{D}_u)\cap\mathcal{D}_u\right)\cup\left(\{\mathcal{N}(\bm{x},y)|(\bm{x},y)\in d^{(t)}\cap\mathcal{D}_u\}\cap\mathcal{D}_u\right) \\
&=\{\mathrm{argmin}_{(\bm{x}^\prime,y^\prime)\in\mathcal{D}\backslash\mathcal{D}_u,y^\prime=y}\|\bm{x}^\prime-\bm{x}\||(\bm{x},y)\in d^{(t)}\cap\mathcal{D}_u\}\cap\mathcal{D}_u \\
&=\emptyset.
\end{aligned}
\end{equation}
The last equality sign holds because each element $\mathrm{argmin}_{(\bm{x}^\prime,y^\prime)\in\mathcal{D}\backslash\mathcal{D}_u,y^\prime=y}\|\bm{x}^\prime-\bm{x}\|$ is involved in $\mathcal{D}\backslash\mathcal{D}_u$.
Hence, we have verified the removable property of \cref{alg:adv2}.
Next, we verify the reproducible property.
For any $t\in\mathcal{I}$, we have $t\in\mathcal{I}_e$ or $t\in\mathcal{I}_n$. We then discuss these two conditions separately.
For simplicity, we denote that $\bm{p}^{(t)}=\bm{w}_r^{(t)}-\bm{w}^{(t)}$.

(1). $t\in\mathcal{I}_e$: we have $d_r^{(t)}=d^{(t)}$.
\begin{equation}
\begin{aligned}
&\bm{w}_r^{(t)}-g_r^{(t)}(\bm{w}_r^{(t-1)},d_r^{(t)}) \\
=&\bm{w}^{(t)}+\bm{p}^{(t)}-\bm{w}_r^{(t-1)}+\gamma^{(t)}\nabla\mathcal{L}(\bm{w}_r^{(t-1)},d_r^{(t)}) \\
=&\bm{w}^{(t)}+\bm{p}^{(t)}-\bm{w}^{(t-1)}-\bm{p}^{(t-1)}+\gamma^{(t)}\nabla\mathcal{L}(\bm{w}_r^{(t-1)},d_r^{(t)}) \\
=&\bm{p}^{(t)}-\bm{p}^{(t-1)}+\gamma^{(t)}\nabla\mathcal{L}(\bm{w}_r^{(t-1)},d^{(t)})-\gamma^{(t)}\nabla\mathcal{L}(\bm{w}^{(t-1)},d^{(t)}). \\
\end{aligned}
\end{equation}
We then focus on the norm value and have
\begin{equation}
\begin{aligned}
&\|\bm{w}_r^{(t)}-g_r^{(t)}(\bm{w}_r^{(t-1)},d_r^{(t)})\| \\
=&\|\bm{p}^{(t)}-\bm{p}^{(t-1)}+\gamma^{(t)}\nabla\mathcal{L}(\bm{w}_r^{(t-1)},d^{(t)})-\gamma^{(t)}\nabla\mathcal{L}(\bm{w}^{(t-1)},d^{(t)})\| \\
\leq&\|\bm{p}^{(t)}-\bm{p}^{(t-1)}\|+\gamma^{(t)}L\|\bm{w}_r^{(t-1)}-\bm{w}^{(t-1)}\| \\
=&\|\bm{p}^{(t)}-\bm{p}^{(t-1)}\|+\gamma^{(t)}L\|\bm{p}^{(t-1)}\|.
\end{aligned}
\end{equation}

(2). $t\in\mathcal{I}_n$: we have $d_r^{(t)}=\mathcal{S}_n(d^{(t)})$.
\begin{equation}
\small
\begin{aligned}
&\bm{w}_r^{(t)}-g_r^{(t)}(\bm{w}_r^{(t-1)},d_r^{(t)}) \\
=&\bm{w}^{(t)}+\bm{p}^{(t)}-\bm{w}_r^{(t-1)}+\gamma^{(t)}\nabla\mathcal{L}(\bm{w}_r^{(t-1)},d_r^{(t)}) \\
=&\bm{w}^{(t-1)}-\gamma^{(t)}\nabla\mathcal{L}(\bm{w}^{(t-1)},d^{(t)})+\bm{p}^{(t)}-\bm{w}_r^{(t-1)}+\gamma^{(t)}\nabla\mathcal{L}(\bm{w}_r^{(t-1)},d_r^{(t)}) \\
=&\bm{p}^{(t)}-\bm{p}^{(t-1)}+\gamma^{(t)}\nabla\mathcal{L}(\bm{w}^{(t-1)},d_r^{(t)})-\gamma^{(t)}\nabla\mathcal{L}(\bm{w}^{(t-1)},d^{(t)})+\gamma^{(t)}\nabla\mathcal{L}(\bm{w}_r^{(t-1)},d_r^{(t)})-\gamma^{(t)}\nabla\mathcal{L}(\bm{w}^{(t-1)},d_r^{(t)}).
\end{aligned}
\end{equation}
We then focus on the norm value and have
\begin{equation}
\small
\begin{aligned}
&\|\bm{w}_r^{(t)}-g_r^{(t)}(\bm{w}_r^{(t-1)},d_r^{(t)})\| \\
\leq&\|\bm{p}^{(t)}-\bm{p}^{(t-1)}\|+\|\gamma^{(t)}\nabla\mathcal{L}(\bm{w}^{(t-1)},d_r^{(t)})-\gamma^{(t)}\nabla\mathcal{L}(\bm{w}^{(t-1)},d^{(t)})\|+\|\gamma^{(t)}\nabla\mathcal{L}(\bm{w}_r^{(t-1)},d_r^{(t)})-\gamma^{(t)}\nabla\mathcal{L}(\bm{w}^{(t-1)},d_r^{(t)})\| \\
\leq&\|\bm{p}^{(t)}-\bm{p}^{(t-1)}\|+\gamma^{(t)}\left(\|\bm{b}^{(t-1)}\|+L\|\bm{p}^{(t-1)}\|\right), \\
\end{aligned}
\end{equation}
where $\bm{b}^{(t-1)}=\nabla\mathcal{L}(\bm{w}^{(t-1)},d_r^{(t)})-\nabla\mathcal{L}(\bm{w}^{(t-1)},d^{(t)})$.
Next, we find the upper bound of $\bm{p}^{(t)}$.
When $t\in\mathcal{I}_e$, we have $\|\bm{p}^{(t)}\|=\gamma_r^{(t)}\|\nabla l(f_{\bm{w}^{(t)}}(\bm{x}^{(t)}),y^{(t)})\|$; when $t\in\mathcal{I}_n$, we have
\begin{equation}
\|\bm{p}^{(t)}\|=\|\bm{w}^{(t-1)}-\gamma^{(t)}\nabla\mathcal{L}(\bm{w}^{(t-1)},d^{(t)})-\bm{w}^{(t-1)}+\gamma^{(t)}\nabla\mathcal{L}(\bm{w}^{(t-1)},d_r^{(t)})\|=\gamma^{(t)}\|\bm{b}^{(t-1)}\|.
\end{equation}
In addition, we know from \cref{eq:bounded bias} that $\|\bm{b}^{(t-1)}\|\leq L_xC_D$.
Combine the case $t\in\mathcal{I}_e$ and the case $t\in\mathcal{I}_n$, we have $\|\bm{p}^{(t)}\|\leq\max\{\gamma_r^{(t)}\|\nabla l(f_{\bm{w}^{(t)}}(\bm{x}^{(t)}),y^{(t)})\|,\gamma^{(t)}L_xC_D\}=P$.
Finally, we verify the reproducible property by proving $\|\bm{p}^{(t)}-\bm{p}^{(t-1)}\|+\gamma^{(t)}L\|\bm{p}^{(t-1)}\|\leq\varepsilon$ and $\|\bm{p}^{(t)}-\bm{p}^{(t-1)}\|+\gamma^{(t)}\left(\|\bm{b}^{(t-1)}\|+L\|\bm{p}^{(t-1)}\|\right)\leq\varepsilon$.
Noting that $\|\bm{p}^{(t)}-\bm{p}^{(t-1)}\|+\gamma^{(t)}\left(\|\bm{b}^{(t-1)}\|+L\|\bm{p}^{(t-1)}\|\right)\geq\|\bm{p}^{(t)}-\bm{p}^{(t-1)}\|+\gamma^{(t)}L\|\bm{p}^{(t-1)}\|$, we can only verify the second inequality.

(i). If $\gamma_r^{(t)}\|\nabla l(f_{\bm{w}^{(t)}}(\bm{x}^{(t)}),y^{(t)})\|\leq\gamma^{(t)}L_xC_D$, we have
\begin{equation}\label{eq:inequality case1}
\|\bm{p}^{(t)}-\bm{p}^{(t-1)}\|+\gamma^{(t)}\left(\|\bm{b}^{(t-1)}\|+L\|\bm{p}^{(t-1)}\|\right)\leq3\gamma^{(t)}L_xC_D+\left(\gamma^{(t)}\right)^2LL_xC_D\leq\varepsilon.
\end{equation}
By solving this quadratic inequality, we obtain that when $\gamma^{(t)}\leq\frac{1}{2L}\left(\sqrt{9+\frac{4\varepsilon L}{L_xC_D}}-3\right)$, \cref{eq:inequality case1} holds.

(ii). If $\gamma_r^{(t)}\|\nabla l(f_{\bm{w}^{(t)}}(\bm{x}^{(t)}),y^{(t)})\|\geq\gamma^{(t)}L_xC_D$, we have
\begin{equation}\label{eq:inequality case2}
\|\bm{p}^{(t)}-\bm{p}^{(t-1)}\|+\gamma^{(t)}\left(\|\bm{b}^{(t-1)}\|+L\|\bm{p}^{(t-1)}\|\right)\leq\gamma^{(t)}L_xC_D+\gamma_r^{(t)}\|\nabla l(f_{\bm{w}^{(t)}}(\bm{x}^{(t)}),y^{(t)})\|(2+\gamma^{(t)}L)\leq\varepsilon.
\end{equation}
Consequently, we obtain that when $\gamma_r^{(t)}\leq\frac{\varepsilon-\gamma^{(t)}L_xC_D}{\|\nabla l(f_{\bm{w}^{(t)}}(\bm{x}^{(t)}),y^{(t)})\|(2+\gamma^{(t)}L)}$, \cref{eq:inequality case2} holds.
In order to avoid no solution for $\gamma_r^{(t)}$, we also need
\begin{equation}\label{eq:inequality make sure solution}
0\leq\frac{\gamma^{(t)}L_xC_D}{\|\nabla l(f_{\bm{w}^{(t)}}(\bm{x}^{(t)}),y^{(t)})\|(2+\gamma^{(t)}L)}\leq\gamma_r^{(t)}\leq\frac{\varepsilon-\gamma^{(t)}L_xC_D}{\|\nabla l(f_{\bm{w}^{(t)}}(\bm{x}^{(t)}),y^{(t)})\|(2+\gamma^{(t)}L)}.
\end{equation}
By solving this inequality for $\gamma^{(t)}$, we obtain $0\leq\gamma^{(t)}\leq\frac{1}{2L}\left(\sqrt{9+\frac{4\varepsilon L}{L_xC_D}}-3\right)$ again.
Combining conclusions from above cases and the assumption that $\|\nabla_{\bm{w}}l(\bm{w},\bm{x})\|\leq G$, we have obtained that when $0\leq\gamma^{(t)}\leq\frac{1}{2L}\left(\sqrt{9+\frac{4\varepsilon L}{L_xC_D}}-3\right)$ and $0\leq\gamma_r^{(t)}\leq\frac{\varepsilon-\gamma^{(t)}L_xC_D}{G(2+\gamma^{(t)}L)}$, the reproducible property holds, i.e., $\|\bm{w}_r^{(t)}-g_r^{(t)}(\bm{w}_r^{(t-1)},d_r^{(t)})\|\leq\varepsilon$.
In conclusion, by incorporating the reproducible property and the removable property, our proof is finished.
\end{proof}



\subsection{Proof of \cref{pro:time_complexity}}
\begin{proof}
We assume the time complexity of naive retraining is $T(n)$.
For \cref{alg:adv1}, the only difference with naive retraining is the selection of mini-batches. 
At best, we can compute the nearest neighbor of each data sample beforehand and obtain a $T(n)$ complexity.
For \cref{alg:adv2}, every iteration in the PoRT can be computed separately in parallel. 
In particular, we have $T(n)=n\cdot m\cdot T(1)$, where $n$ is the number of iterations, $m$ is the batch size, and $T(1)$ denotes the time for computing the gradient of one data sample.
If $t\in\mathcal{I}_e$, computing $\bm{w}_r^{(t)}$ requires a complexity of $T(1)$; if $t\in\mathcal{I}_n$, computing $\bm{w}_r^{(t)}$ requires a complexity of $m\cdot T(1)$.
According to the analysis in \cref{sec:converge}, we have that when $n\rightarrow\infty$, the number of iterations in $\mathcal{I}_n$ will approach $p_un$ and the number of iterations in $\mathcal{I}_e$ will approach $(1-p_u)n$.
Finally, with the parallel processes, we have the overall complexity is
\begin{equation}
m\cdot T(1)\cdot\frac{p_un}{P}+T(1)\cdot\frac{(1-p_u)n}{P}=\left(\frac{p_u}{P}+\frac{1-p_u}{mP}\right)T(n).
\end{equation}
\end{proof}

\section{Experimental Setups and Additional Experimental Results}\label{sec:experiment}


\begin{table}[t]
\scriptsize
\renewcommand{\arraystretch}{1.1}
\tabcolsep = 3pt
\centering
\caption{\textbf{Model Utility in Normal Settings}. We assess the model utility among original training, naive retraining, and adversarial unlearning methods over three popular DNNs across three real-world datasets. We record the macro F1-score of the predictions on the unlearned set $\mathcal{D}_u$, retained set $\mathcal{D}\backslash\mathcal{D}_u$, and test set $\mathcal{D}_t$.}
\label{tab:normal_full}
\aboverulesep = 0pt
\belowrulesep = 0pt
\begin{tabular}{c|c|ccc|ccc|ccc}
\toprule
\multirow{2}{*}{\textbf{Model}} & \multirow{2}{*}{\textbf{Method}} & \multicolumn{3}{c|}{\textbf{MNIST}} & \multicolumn{3}{c|}{\textbf{CIFAR-10}} & \multicolumn{3}{c}{\textbf{SVHN}} \\
& & $\mathcal{D}_u$ & $\mathcal{D}\backslash\mathcal{D}_u$ & $\mathcal{D}_t$ & $\mathcal{D}_u$ & $\mathcal{D}\backslash\mathcal{D}_u$ & $\mathcal{D}_t$ & $\mathcal{D}_u$ & $\mathcal{D}\backslash\mathcal{D}_u$ & $\mathcal{D}_t$ \\
\midrule
\multirow{5}{*}{MLP} & Original & 99.47 $\pm$ 0.09 & 99.76 $\pm$ 0.08 & 97.00 $\pm$ 0.17 & 80.96 $\pm$ 0.15 & 80.75 $\pm$ 0.96 & 50.54 $\pm$ 0.57 & 89.56 $\pm$ 0.78 & 89.40 $\pm$ 0.45 & 80.03 $\pm$ 0.45 \\
& Retrain & 96.43 $\pm$ 0.19 & 99.52 $\pm$ 0.10 & 96.75 $\pm$ 0.13 & 49.32 $\pm$ 0.59 & 82.96 $\pm$ 1.29 & 49.29 $\pm$ 0.77 & 82.42 $\pm$ 0.22 & 89.83 $\pm$ 0.39 & 79.50 $\pm$ 0.37 \\
& Adv-R ($\mathcal{S}_r$) & 96.36 $\pm$ 0.13 & 99.48 $\pm$ 0.17 & 96.58 $\pm$ 0.12 & 48.49 $\pm$ 0.41 & 82.33 $\pm$ 0.40 & 48.27 $\pm$ 0.48 & 81.85 $\pm$ 0.85 & 89.31 $\pm$ 0.86 & 78.63 $\pm$ 0.71 \\
& Adv-R ($\mathcal{S}_n$) & 96.34 $\pm$ 0.11 & 98.65 $\pm$ 0.19 & 96.60 $\pm$ 0.14 & 50.43 $\pm$ 0.17 & 80.01 $\pm$ 1.30 & 49.86 $\pm$ 0.23 & 82.98 $\pm$ 0.57 & 92.60 $\pm$ 0.56 & 79.77 $\pm$ 0.57 \\
& Adv-F & 99.30 $\pm$ 0.13 & 99.33 $\pm$ 0.10 & 96.94 $\pm$ 0.14 & 81.13 $\pm$ 0.76 & 81.38 $\pm$ 0.79 & 50.81 $\pm$ 0.63 & 89.99 $\pm$ 0.71 & 89.77 $\pm$ 0.55 & 80.11 $\pm$ 0.85 \\
\midrule
\multirow{5}{*}{CNN} & Original & 99.69 $\pm$ 0.31 & 99.74 $\pm$ 0.30 & 99.13 $\pm$ 0.25 & 100.00 $\pm$ 0.00 & 100.00 $\pm$ 0.00 & 85.33 $\pm$ 0.31 & 99.64 $\pm$ 0.73 & 99.61 $\pm$ 0.77 & 94.66 $\pm$ 1.00 \\
& Retrain & 99.31 $\pm$ 0.12 & 100.00 $\pm$ 0.00 & 99.39 $\pm$ 0.05 & 83.60 $\pm$ 0.31 & 100.00 $\pm$ 0.00 & 83.12 $\pm$ 0.23 & 94.24 $\pm$ 0.22 & 100.00 $\pm$ 0.00 & 94.96 $\pm$ 0.10 \\
& Adv-R ($\mathcal{S}_r$) & 99.24 $\pm$ 0.13 & 100.00 $\pm$ 0.00 & 99.27 $\pm$ 0.02 & 83.81 $\pm$ 0.44 & 100.00 $\pm$ 0.00 & 83.08 $\pm$ 0.34 & 94.29 $\pm$ 0.17 & 100.00 $\pm$ 0.00 & 94.90 $\pm$ 0.07 \\
& Adv-R ($\mathcal{S}_n$) & 99.35 $\pm$ 0.09 & 100.00 $\pm$ 0.00 & 99.40 $\pm$ 0.02 & 83.12 $\pm$ 0.31 & 100.00 $\pm$ 0.00 & 82.71 $\pm$ 0.33 & 94.36 $\pm$ 0.32 & 100.00 $\pm$ 0.00 & 94.92 $\pm$ 0.15 \\
& Adv-F & 100.00 $\pm$ 0.00 & 100.00 $\pm$ 0.00 & 99.46 $\pm$ 0.08 & 100.00 $\pm$ 0.00 & 100.00 $\pm$ 0.00 & 85.20 $\pm$ 0.24 & 100.00 $\pm$ 0.00 & 99.99 $\pm$ 0.01 & 95.13 $\pm$ 0.10 \\
\midrule
\multirow{5}{*}{ResNet} & Original & 100.00 $\pm$ 0.00 & 100.00 $\pm$ 0.00 & 99.53 $\pm$ 0.05 & 99.99 $\pm$ 0.03 & 99.99 $\pm$ 0.03 & 84.11 $\pm$ 0.74 & 100.00 $\pm$ 0.00 & 100.00 $\pm$ 0.00 & 94.91 $\pm$ 0.09 \\
& Retrain & 99.41 $\pm$ 0.09 & 100.00 $\pm$ 0.00 & 99.46 $\pm$ 0.07 & 82.76 $\pm$ 0.38 & 100.00 $\pm$ 0.00 & 82.10 $\pm$ 0.39 & 94.33 $\pm$ 0.24 & 100.00 $\pm$ 0.00 & 94.57 $\pm$ 0.06 \\
& Adv-R ($\mathcal{S}_r$) & 99.43 $\pm$ 0.09 & 100.00 $\pm$ 0.00 & 99.46 $\pm$ 0.04 & 82.29 $\pm$ 0.98 & 99.99 $\pm$ 0.01 & 81.71 $\pm$ 0.78 & 94.38 $\pm$ 0.11 & 100.00 $\pm$ 0.00 & 94.54 $\pm$ 0.09 \\
& Adv-R ($\mathcal{S}_n$) & 99.41 $\pm$ 0.06 & 100.00 $\pm$ 0.00 & 99.42 $\pm$ 0.04 & 82.40 $\pm$ 0.39 & 100.00 $\pm$ 0.00 & 81.85 $\pm$ 0.44 & 94.64 $\pm$ 0.20 & 100.00 $\pm$ 0.00 & 94.75 $\pm$ 0.04 \\
& Adv-F & 100.00 $\pm$ 0.00 & 100.00 $\pm$ 0.00 & 99.49 $\pm$ 0.03 & 100.00 $\pm$ 0.00 & 100.00 $\pm$ 0.00 & 84.54 $\pm$ 0.37 & 100.00 $\pm$ 0.00 & 100.00 $\pm$ 0.00 & 94.91 $\pm$ 0.07 \\
\bottomrule
\end{tabular}
\end{table}

\begin{table}[t]
\scriptsize
\renewcommand{\arraystretch}{1.1}
\tabcolsep = 3pt
\centering
\caption{\textbf{Model Utility in Class-Imbalanced Settings}. We assess the model utility among original training, naive retraining, and adversarial unlearning methods over three popular DNNs across three real-world datasets. We record the macro F1-score of the predictions on the unlearned set $\mathcal{D}_u$, retained set $\mathcal{D}\backslash\mathcal{D}_u$, and test set $\mathcal{D}_t$.}
\label{tab:imbalance_full}
\aboverulesep = 0pt
\belowrulesep = 0pt
\begin{tabular}{c|c|ccc|ccc|ccc}
\toprule
\multirow{2}{*}{\textbf{Model}} & \multirow{2}{*}{\textbf{Method}} & \multicolumn{3}{c|}{\textbf{MNIST}} & \multicolumn{3}{c|}{\textbf{CIFAR-10}} & \multicolumn{3}{c}{\textbf{SVHN}} \\
& & $\mathcal{D}_u$ & $\mathcal{D}\backslash\mathcal{D}_u$ & $\mathcal{D}_t$ & $\mathcal{D}_u$ & $\mathcal{D}\backslash\mathcal{D}_u$ & $\mathcal{D}_t$ & $\mathcal{D}_u$ & $\mathcal{D}\backslash\mathcal{D}_u$ & $\mathcal{D}_t$ \\
\midrule
\multirow{5}{*}{MLP} & Original & 60.29 $\pm$ 13.07 & 97.00 $\pm$ 2.60 & 96.88 $\pm$ 0.07 & 34.50 $\pm$ 6.41 & 75.68 $\pm$ 2.29 & 50.95 $\pm$ 0.19 & 39.13 $\pm$ 7.26 & 84.02 $\pm$ 3.90 & 80.50 $\pm$ 0.83 \\
& Retrain & 38.76 $\pm$ 13.41 & 95.86 $\pm$ 4.34 & 89.92 $\pm$ 5.70 & 13.29 $\pm$ 4.49 & 77.93 $\pm$ 2.83 & 44.81 $\pm$ 1.70 & 22.65 $\pm$ 8.11 & 76.87 $\pm$ 3.93 & 67.33 $\pm$ 4.90 \\
& Adv-R ($\mathcal{S}_r$) & 39.48 $\pm$ 12.20 & 99.71 $\pm$ 0.19 & 91.04 $\pm$ 4.80 & 13.13 $\pm$ 4.32 & 81.21 $\pm$ 3.21 & 44.19 $\pm$ 1.81 & 26.33 $\pm$ 8.99 & 87.25 $\pm$ 2.74 & 72.09 $\pm$ 4.50 \\
& Adv-R ($\mathcal{S}_n$) & 42.90 $\pm$ 11.87 & 97.96 $\pm$ 0.59 & 92.80 $\pm$ 4.54 & 16.17 $\pm$ 4.76 & 67.25 $\pm$ 2.15 & 45.74 $\pm$ 1.80 & 28.06 $\pm$ 9.45 & 82.16 $\pm$ 1.65 & 70.67 $\pm$ 3.55 \\
& Adv-F & 64.21 $\pm$ 9.89 & 97.28 $\pm$ 3.34 & 96.81 $\pm$ 0.04 & 35.46 $\pm$ 6.74 & 75.08 $\pm$ 1.68 & 51.06 $\pm$ 0.86 & 40.10 $\pm$ 7.87 & 82.78 $\pm$ 4.54 & 79.70 $\pm$ 1.08 \\
\midrule
\multirow{5}{*}{CNN} & Original & 100.00 $\pm$ 0.00 & 100.00 $\pm$ 0.00 & 99.48 $\pm$ 0.06 & 100.00 $\pm$ 0.00 & 100.00 $\pm$ 0.00 & 85.44 $\pm$ 0.22 & 100.00 $\pm$ 0.00 & 100.00 $\pm$ 0.00 & 95.13 $\pm$ 0.07 \\
& Retrain & 47.64 $\pm$ 15.76 & 100.00 $\pm$ 0.00 & 94.98 $\pm$ 5.35 & 24.25 $\pm$ 6.98 & 90.88 $\pm$ 5.03 & 65.22 $\pm$ 5.94 & 32.79 $\pm$ 12.61 & 92.68 $\pm$ 4.79 & 84.23 $\pm$ 6.65 \\
& Adv-R ($\mathcal{S}_r$) & 46.00 $\pm$ 12.84 & 100.00 $\pm$ 0.00 & 94.94 $\pm$ 4.77 & 25.94 $\pm$ 6.89 & 96.00 $\pm$ 4.90 & 76.51 $\pm$ 4.32 & 33.43 $\pm$ 12.12 & 97.52 $\pm$ 3.87 & 86.69 $\pm$ 5.40 \\
& Adv-R ($\mathcal{S}_n$) & 49.90 $\pm$ 15.03 & 99.96 $\pm$ 0.07 & 95.08 $\pm$ 4.82 & 23.97 $\pm$ 8.75 & 91.46 $\pm$ 1.81 & 67.34 $\pm$ 4.02 & 33.89 $\pm$ 12.26 & 98.49 $\pm$ 0.40 & 85.76 $\pm$ 6.46 \\
& Adv-F & 92.49 $\pm$ 15.01 & 99.97 $\pm$ 0.05 & 99.45 $\pm$ 0.12 & 100.00 $\pm$ 0.00 & 100.00 $\pm$ 0.00 & 85.11 $\pm$ 0.21 & 100.00 $\pm$ 0.00 & 100.00 $\pm$ 0.00 & 95.14 $\pm$ 0.05 \\
\midrule
\multirow{5}{*}{ResNet} & Original & 100.00 $\pm$ 0.00 & 100.00 $\pm$ 0.00 & 99.54 $\pm$ 0.05 & 98.30 $\pm$ 3.39 & 99.96 $\pm$ 0.08 & 85.02 $\pm$ 0.96 & 100.00 $\pm$ 0.00 & 100.00 $\pm$ 0.00 & 94.66 $\pm$ 0.30 \\
& Retrain & 50.74 $\pm$ 14.51 & 99.00 $\pm$ 2.00 & 95.89 $\pm$ 5.40 & 21.80 $\pm$ 6.43 & 89.13 $\pm$ 2.84 & 66.86 $\pm$ 3.29 & 33.08 $\pm$ 11.97 & 95.19 $\pm$ 3.78 & 83.89 $\pm$ 5.83 \\
& Adv-R ($\mathcal{S}_r$) & 51.76 $\pm$ 14.11 & 100.00 $\pm$ 0.00 & 96.33 $\pm$ 4.73 & 25.53 $\pm$ 6.93 & 100.00 $\pm$ 0.00 & 74.18 $\pm$ 4.26 & 34.76 $\pm$ 12.06 & 98.00 $\pm$ 4.01 & 87.63 $\pm$ 6.11 \\
& Adv-R ($\mathcal{S}_n$) & 50.61 $\pm$ 11.18 & 100.00 $\pm$ 0.00 & 96.54 $\pm$ 4.27 & 23.83 $\pm$ 6.84 & 96.53 $\pm$ 1.05 & 68.86 $\pm$ 4.00 & 33.92 $\pm$ 11.85 & 99.37 $\pm$ 0.20 & 84.87 $\pm$ 5.42 \\
& Adv-F & 100.00 $\pm$ 0.00 & 100.00 $\pm$ 0.00 & 99.54 $\pm$ 0.03 & 97.13 $\pm$ 5.74 & 99.98 $\pm$ 0.03 & 84.04 $\pm$ 0.46 & 100.00 $\pm$ 0.00 & 100.00 $\pm$ 0.00 & 94.77 $\pm$ 0.09 \\
\bottomrule
\end{tabular}
\end{table}

\begin{table}[t]
\renewcommand{\arraystretch}{1.05}
\centering
\caption{Comparison of the model utility among original training, naive retraining, and adversarial unlearning methods based on ResNet-50 over the Tiny-ImageNet dataset. We record the macro F1-score of the predictions on the unlearned set $\mathcal{D}_u$, retained set $\mathcal{D}\backslash\mathcal{D}_u$, and test set $\mathcal{D}_t$.}
\label{tab:additional utility}
\aboverulesep = 0pt
\belowrulesep = 0pt
\begin{tabular}{c|ccc|ccc}
\toprule
\multirow{2}{*}{\textbf{Method}} & \multicolumn{3}{c|}{\textbf{Normal}} & \multicolumn{3}{c}{\textbf{Imbalanced}} \\
& $\mathcal{D}_u$ & $\mathcal{D}\backslash\mathcal{D}_u$ & $\mathcal{D}_t$ & $\mathcal{D}_u$ & $\mathcal{D}\backslash\mathcal{D}_u$ & $\mathcal{D}_t$ \\
\midrule
Original & 90.34 $\pm$ 0.47 & 90.21 $\pm$ 0.43 & 36.59 $\pm$ 0.74 & 91.76 $\pm$ 0.64 & 91.42 $\pm$ 0.59 & 33.81 $\pm$ 0.60 \\
Retrain & 31.03 $\pm$ 0.62 & 93.78 $\pm$ 0.57 & 31.08 $\pm$ 0.33 & 8.59 $\pm$ 1.94 & 95.79 $\pm$ 18.38 & 26.19 $\pm$ 1.91 \\
Adv-R ($\mathcal{S}_r$) & 31.26 $\pm$ 0.46 & 92.83 $\pm$ 0.63 & 31.78 $\pm$ 0.16 & 8.98 $\pm$ 2.95 & 95.65 $\pm$ 18.65 & 26.76 $\pm$ 1.17 \\
Adv-R ($\mathcal{S}_n$) & 31.82 $\pm$ 0.26 & 93.09 $\pm$ 0.42 & 31.76 $\pm$ 0.43 & 8.70 $\pm$ 0.96 & 92.83 $\pm$ 5.36 & 26.80 $\pm$ 1.20 \\
Adv-F & 90.16 $\pm$ 0.66 & 90.30 $\pm$ 0.51 & 36.57 $\pm$ 0.80 & 91.37 $\pm$ 0.56 & 91.10 $\pm$ 0.56 & 33.69 $\pm$ 0.81 \\
\bottomrule
\end{tabular}
\end{table}



We implemented all experiments in the PyTorch~\citep{paszke2019pytorch} library and exploited SGD as the optimizer for training.
For consistent hyperparameter settings across all datasets, we fix the learning rate $\gamma^{(t)}$ as $10^{-2}$, the weight decay parameter as $5\times10^{-4}$, the training epochs number as $30$, and set the batch size to 128. In determining the selection $\mathcal{S}_r$, specifically for data batches containing unlearned data, we randomly sample $50$ data batches from the remaining set $\mathcal{D}\backslash\mathcal{D}_u$ and select the batch that yields the smallest distance, as defined in~\cref{equ:random}. All experiments were conducted on an Nvidia RTX A6000 GPU. We reported the average value and the standard deviation of the numerical results under five different random seeds. 
For the experiment of verification errors in \cref{fig:verify_error}, we set the learning rate $\gamma^{(t)}$ as $5\times10^{-3}$, weight decay parameter as $0$.
We fix the size of the unlearned set as $1,000$ and set the learning rate $\gamma_r^{(t)}$ as $10^{-3}$.

\subsection{Model Utility Study}\label{sec:utl_exp}
\subsubsection{Experimental Settings}
All datasets utilized in our experiments adhere to the standard train/test split provided by the Torchvision library~\cite{torchvision2016}. Within each experiment, $20\%$ of the training data is set aside as the validation set. To assess model performance, we randomly sample $10\%$ of the remaining training data to form the unlearn set. We report the average macro F1-score, along with its standard deviation, based on model predictions for the unlearn set $\mathcal{D}_u$, the remaining set $\mathcal{D}\backslash\mathcal{D}_u$, and the test set $\mathcal{D}_t$. These metrics are computed across five random seeds to ensure robustness.

Additionally, to simulate scenarios where the data distribution of the unlearn set deviates from the overall training set, we introduce a class-imbalanced unlearning setting under a fixed train/val/test split. For each class $c$, we follow the approach of~\citeauthor{yurochkin2019bayesian} by drawing a 5-dimensional vector $q_c$ from a Dirichlet distribution with its parameter of 0.5. We then assign data to the $i$-th piece proportionally to $q_c[i]$. In each experiment, one piece of data is selected as the unlearn set $\mathcal{D}_u$, while the remaining pieces constitute the remaining set $\mathcal{D}\backslash\mathcal{D}_u$. The average model performance is recorded across five experiments.

\subsubsection{Additional Experimental Results}
We provide the complete experimental results of the model utility in different types of unlearning strategies under normal and class-imbalanced settings in \cref{tab:normal_full} and \cref{tab:imbalance_full}, respectively.
From the full version of the results, we can obtain that: 
\begin{enumerate}
\item The naive retraining can render a larger utility drop when the model under-fits the data and the data heterogeneity exists in the unlearning process;
\item In the normal setting, the retraining-based adversarial method has similar utility to the naive retraining, while in the class-imbalanced setting, the retraining-based adversarial method achieves a much better utility than naive retraining, which means model providers can benefit more when data heterogeneity exists in the unlearning process;
\item The forging-based adversarial method has much better performance than other unlearning methods in all cases, even better than the original training in some cases, which means the model provider can benefit the most from forging (but also with a larger probability of being detected by backdoor verification).
\end{enumerate}

To further show the scalability of our proposed methods, we conduct additional experiments using the ResNet-50 model~\citep{he2016deep} over the Tiny-ImageNet dataset~\citep{le2015tiny}.
The TinyImageNet dataset~\citep{le2015tiny} is a subset of the ImageNet dataset. 
It consists of 200 object classes, and for each object class, it provides 500 training images, 50 validation images, and 50 test images. 
All images have been downsampled to 64 $\times$ 64 $\times$ 3 pixels. 
As the test set is released without labels, we use the validation set as the test set in our experiment. 
Within each experiment, 20\% of the training data is set aside as the validation set, and the division of the unlearn set and other parameter settings are consistent with the main experiments in the paper. 
We train the ResNet-50 model from scratch for 50 epochs following the experimental setting in \citep{yuan2020revisiting} and record the accuracy under 5 random seeds. 
For $S_r$, we randomly sample 10 mini-batches from the retained set $D\backslash D_u$ and select the batch that yields the smallest distance, as defined in \Cref{equ:random}. 
The experimental results of the utility of the unlearned model are shown in \Cref{tab:additional utility}.
Basically, the experimental results are consistent with \Cref{tab:utility} in our paper (though overfitting exists), i.e., Adv-F achieves comparable performance as the original model and Adv-R has a better performance compared with naive retraining. 
We also find that the difference between the normal setting and the imbalanced setting is not as distinct as in \Cref{tab:utility}. 
We explain this as 
\begin{enumerate}
\item Tiny-ImageNet has 200 classes while the datasets in \Cref{tab:utility} only have 10 classes. The impact of imbalanced sampling can be weakened under far more classes.
\item Tiny-ImageNet has only 400 training samples per class while the datasets in \Cref{tab:utility} have over 5,000 samples. The effect of nearest neighbor selection can be diminished as the constant $C_D$ (introduced in \Cref{pro:cvg}) might increase and subsequently, the gap between the adversarially unlearned model and the original model can be larger according to \Cref{pro:cvg}.
\end{enumerate}

\subsection{Backdoor Verification}
\begin{table}[t]
\scriptsize
\renewcommand{\arraystretch}{1.1}
\tabcolsep = 3pt
\centering
\caption{Results of backdoor verification on different (adversarial) unlearning strategies over three datasets.}
\label{tab:backdoor_full}
\aboverulesep = 0pt
\belowrulesep = 0pt
\begin{tabular}{c|ccc|ccc|ccc}
\toprule
\multirow{2}{*}{\textbf{Method}} & \multicolumn{3}{c|}{\textbf{MLP \& MNIST}} & \multicolumn{3}{c|}{\textbf{CNN \& CIFAR-10}} & \multicolumn{3}{c}{\textbf{ResNet \& SVHN}} \\
& in atk acc $p$ & ex atk acc $q$ & $\beta$ & in atk acc $p$ & ex atk acc $q$ & $\beta$ & in atk acc $p$ & ex atk acc $q$ & $\beta$ \\
\midrule
Original & 0.998 $\pm$ 0.007 & 0.101 $\pm$ 0.010 & $2.61\times10^{-42}$ & 0.933 $\pm$ 0.105 & 0.088 $\pm$ 0.139 & $5.14\times10^{-20}$ & 0.982 $\pm$ 0.003 & 0.095 $\pm$ 0.001 & $1.11\times10^{-28}$ \\
Retrain & 0.102 $\pm$ 0.016 & 0.103 $\pm$ 0.012 & 0.998 & 0.118 $\pm$ 0.022 & 0.124 $\pm$ 0.010 & 0.998 & 0.110 $\pm$ 0.001 & 0.096 $\pm$ 0.001 & 0.999\\
Adv-R & 0.103 $\pm$ 0.017 & 0.102 $\pm$ 0.015 & 0.997 & 0.129 $\pm$ 0.021 & 0.102 $\pm$ 0.009 & 0.997 & 0.109 $\pm$ 0.001 & 0.096 $\pm$ 0.001 & 0.999\\
Adv-F & 0.995 $\pm$ 0.003 & 0.103 $\pm$ 0.013 & $5.46\times10^{-34}$ & 0.914 $\pm$ 0.119 & 0.100 $\pm$ 0.050 & $2.78\times10^{-16}$ & 0.981 $\pm$ 0.006 & 0.096 $\pm$ 0.001 & $2.08\times10^{-26}$ \\
\bottomrule
\end{tabular}
\end{table}
For backdoor verification, we exploit Athena~\citep{sommer2022athena} as the verification strategy. Specifically, we consider two hypotheses: $H_0$ - \textit{the data has been unlearned}, and $H_1$ - \textit{the data has not been unlearned}. In assessing the effectiveness of a backdoor verification strategy applied to an algorithm  $\mathcal{A}$, we focus on the deletion confidence for a given acceptable tolerance of Type I error $\alpha$ ( $\alpha =$ Pr[Reject $H_0|H_0$ is true]), i.e.,
\begin{equation}
    \rho_{\mathcal{A},\alpha}(n)=(1-\beta) =1-\text{Pr[Accept}\hspace{0.2em}H_0 |H_1\hspace{0.2em}\text{is true]}.
\end{equation}
We follow \citeauthor{sommer2022athena} to compute the Type II error $\beta$ as a function of $\alpha$ and the number of testing samples $n$, i.e.,
\begin{equation}
    (1-\beta) = 1-\sum_{k=0}^n \tbinom{n}{k}p^k(1-p)^{n-k}\cdot I[\sum_{l=0}^k \tbinom{n}{l}q^l(1-q)^{n-l}\leq 1-\alpha],\label{eqn:beta}
\end{equation}
where $q$ and $p$ represent the backdoor attack accuracy for deleted and undeleted data, respectively. The function $I(x)= 1$ if $x$ is True and 0 otherwise. We apply similar method as in \citeauthor{sommer2022athena} to estimate $q$ and $p$:
\begin{itemize}
    \item \textbf{Estimating $p$}. To estimate $p$, we first introduce a specific backdoor pattern to $10\%$ of the unlearn set. This involves randomly selecting four pixels in each sample, setting their values to 1, and assigning a target label $c_{b}$. The model is then trained on this partially backdoored dataset $\mathcal{D}_{train}^{b}$. For evaluation, we extract $2\%$ of the test samples $\mathcal{D}_{test}^{b}$, apply the same backdoor pattern, and calculate the backdoor success rate for this trigger with the target label $c_{b}$ as,
    \begin{equation}
        p = \frac{1}{|\mathcal{D}_{test}^{b}|}|\{(x,y)\in\mathcal{D}_{test}^{b}|\mathcal{A}(\mathcal{D}_{train}^{b})(x) = c_b \}|
    \end{equation}
    \item \textbf{Estimating $q$}. To estimate $q$, we randomly select an additional $2\%$ of the test samples, denoted as $\mathcal{D}_{test}^{ex}$. On these samples, a different backdoor pattern is applied. We then calculate the backdoor success rate, focusing on the trigger with an alternate target label $c_{ex}$. This process enables us to determine $q$ as,
    \begin{equation}
        q = \frac{1}{|\mathcal{D}_{test}^{ex}|}|\{(x,y)\in\mathcal{D}_{test}^{ex}|\mathcal{A}(\mathcal{D}_{train}^{b})(x) = c_{ex} \}|
    \end{equation}
\end{itemize}
We set $\alpha$ to $10^{-3}$, $n$ to $30$, and use the estimate $p$ and $q$ to compute $\beta$ in~\cref{eqn:beta}. For the MNIST and CIFAR-10 datasets, the target labels $c_{b}$ and $c_{ex}$ are selected randomly, owing to the even distribution of their test data across classes. In contrast, for the SVHN dataset, we specifically choose $c_{b} = 3$ and $c_{ex} = 4$ for a better explanation due to its uneven data distribution. Notably, in the SVHN dataset, data labeled as 3 or 4 accounts for nearly one-tenth of the test data, which is significant given that the dataset comprises ten classes. We provide the complete experimental results of the backdoor verification in \cref{tab:backdoor_full}.

We first clarify that if $p$ or $q$ approaches the level of random prediction, it suggests the corresponding backdoored data has not been utilized during the training process. Conversely, a high value of $p$ indicates the effectiveness of the backdoor strategy. In other words, a large $p$ value signifies that the backdoor attack was successful in influencing the model's behavior. The results in Table~\ref{tab:backdoor_full} reveal that the backdoor verification mechanism is highly effective for models trained on $\mathcal{D}_{train}^{b}$. This effectiveness is primarily attributed to the substantial difference between $p$ and $q$. Additionally, the verification mechanism is capable of detecting models modified using the forging-based method with a high degree of probability. This is due to the fact that such modifications only slightly alter the original model. In contrast, the retraining-based model can deception backdoor verification, as it does not directly utilize the partially backdoored unlearning data, resulting in a model that is not affected by poisoned data.

\begin{table}[t]
    \centering
    \renewcommand{\arraystretch}{1.05}
    \caption{Results of the membership inference attack on different (adversarial) unlearning strategies over three datasets.}\label{tab:membership inference}
    \tabcolsep = 8pt
    \aboverulesep = 0pt
    \belowrulesep = 0pt
    \begin{tabular}{l|ccc}
    \toprule
    \textbf{Method} & \textbf{MNIST} & \textbf{CIFAR-10} & \textbf{SVHN} \\
    \midrule
    Original & 50.60 $\pm$ 0.61 & 72.59 $\pm$ 1.03 & 59.67 $\pm$ 0.31 \\
    Retrain & 50.16 $\pm$ 0.92 & 48.97 $\pm$ 0.65 & 51.88 $\pm$ 1.39 \\
    Adv-F & 50.09 $\pm$ 0.34 & 72.41 $\pm$ 0.55 & 59.04 $\pm$ 1.10 \\
    Adv-R & 49.66 $\pm$ 0.85 & 48.93 $\pm$ 1.22 & 50.09 $\pm$ 0.65 \\
    \bottomrule
    \end{tabular}
\end{table}

\begin{figure}[t]
    \centering
    \subfigure[Verification error distribution]{
    \includegraphics[width=0.35\linewidth]{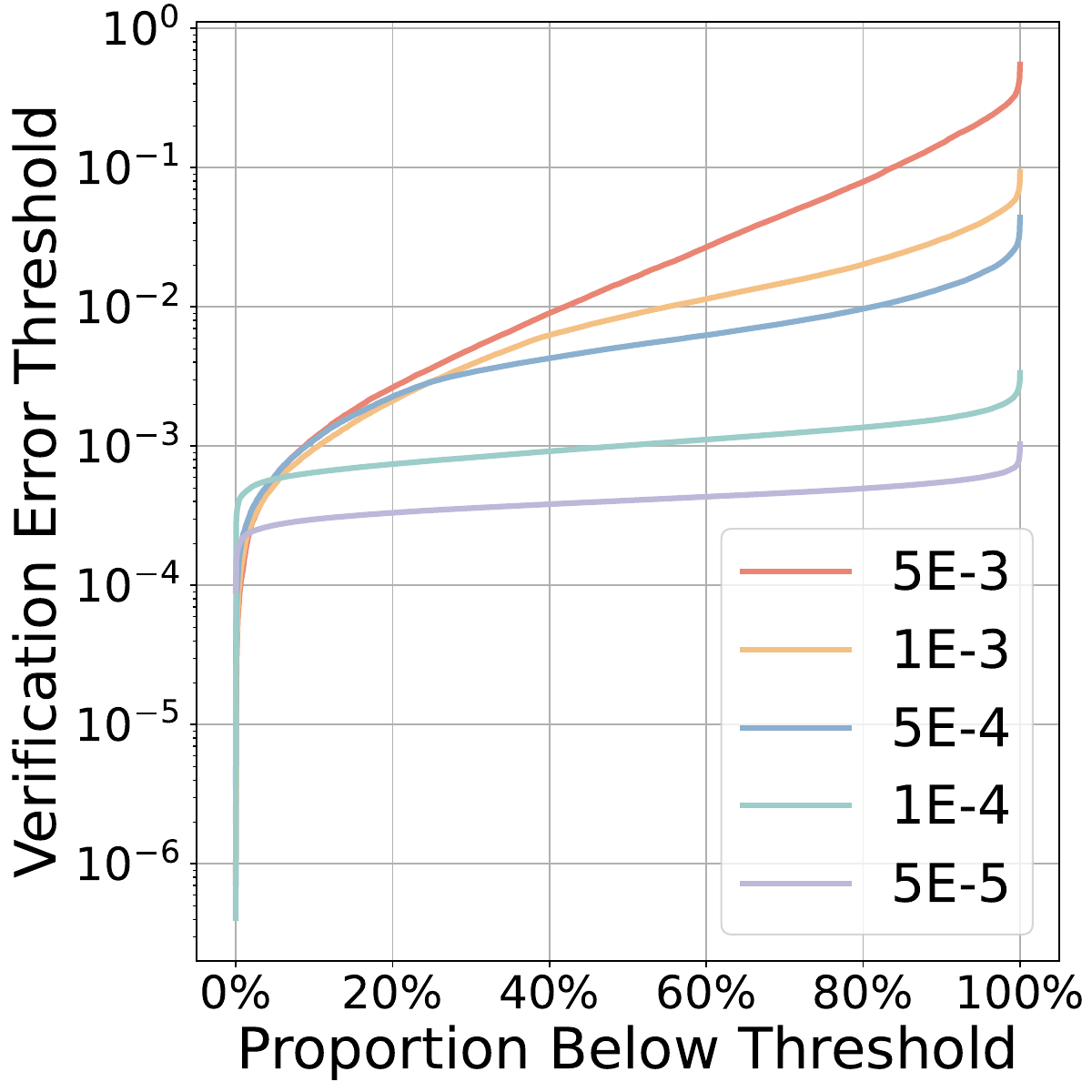}\label{fig:mnist_err}}
    \subfigure[Verification error statistics]{
    \includegraphics[width=0.35\linewidth]{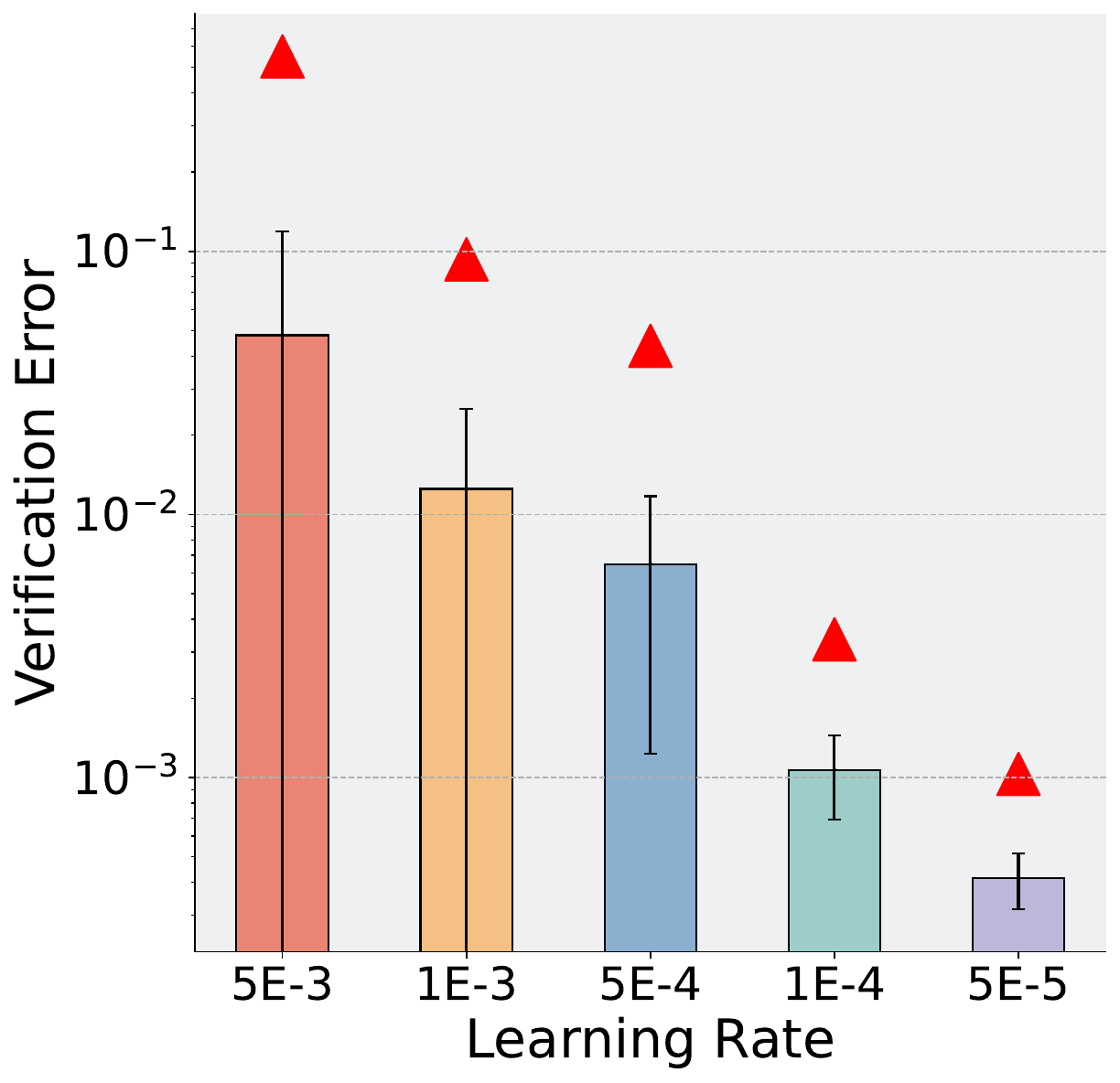}\label{fig:mnist_mean_err}}
    \caption{Comparison of verification error under different learning rate configurations.}
    \label{fig:verification_error_lr}
\end{figure}

\begin{figure}[t]
    \centering
    \subfigure[Gradient norm over MNIST.]{
    \includegraphics[width=0.3\linewidth]{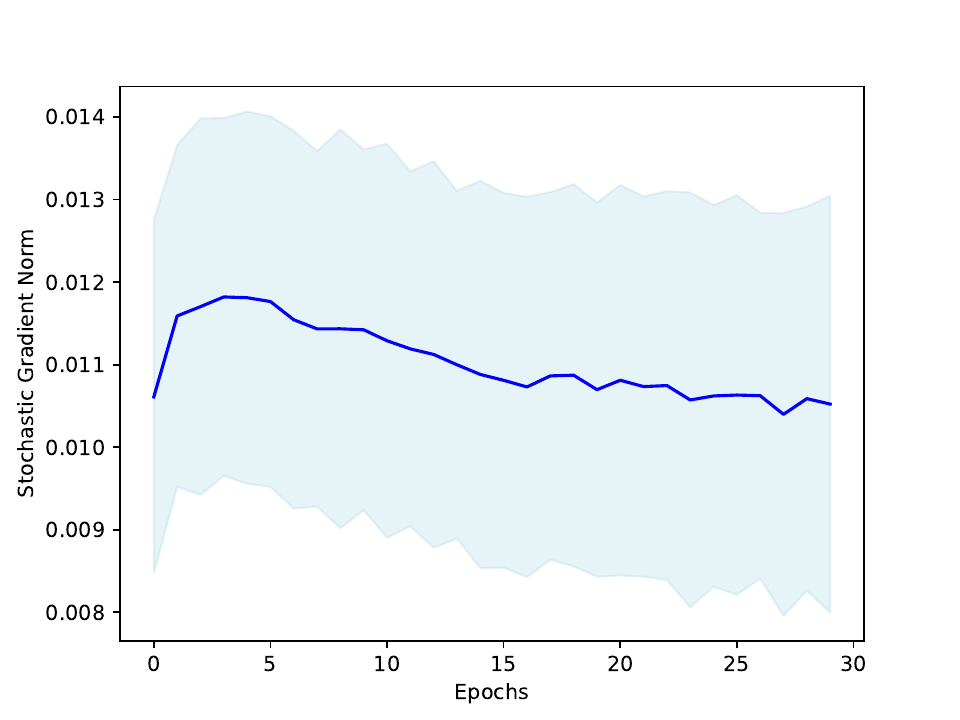}\label{fig:gradient_mnist}}
    \subfigure[Gradient norm over CIFAR-10.]{
    \includegraphics[width=0.3\linewidth]{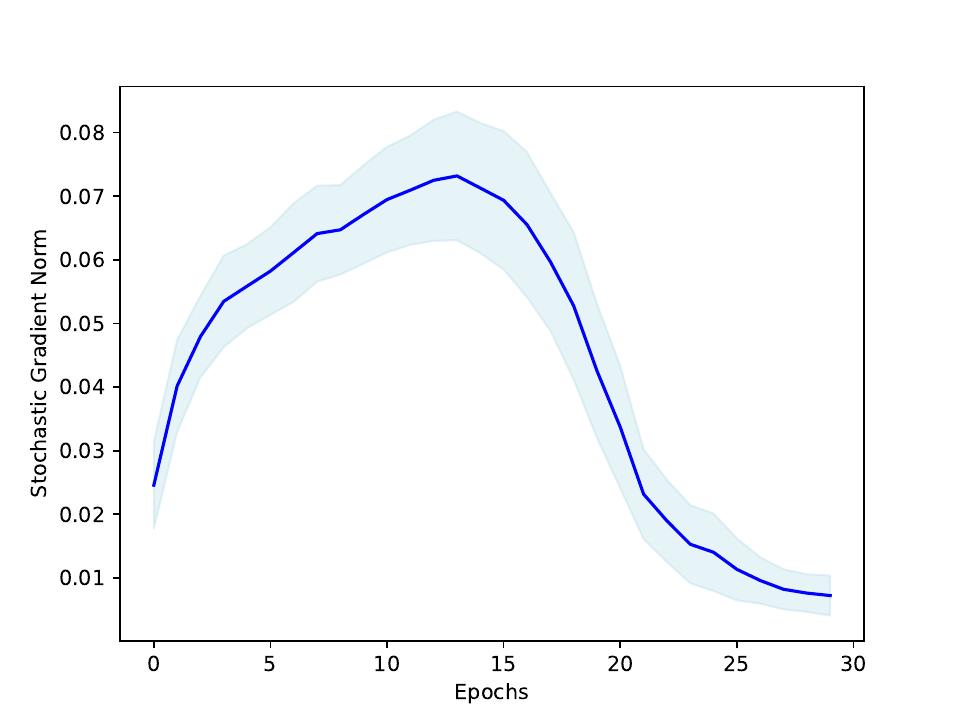}\label{fig:gradient_cifar}}
    \subfigure[Gradient norm over SVHN.]{
    \includegraphics[width=0.3\linewidth]{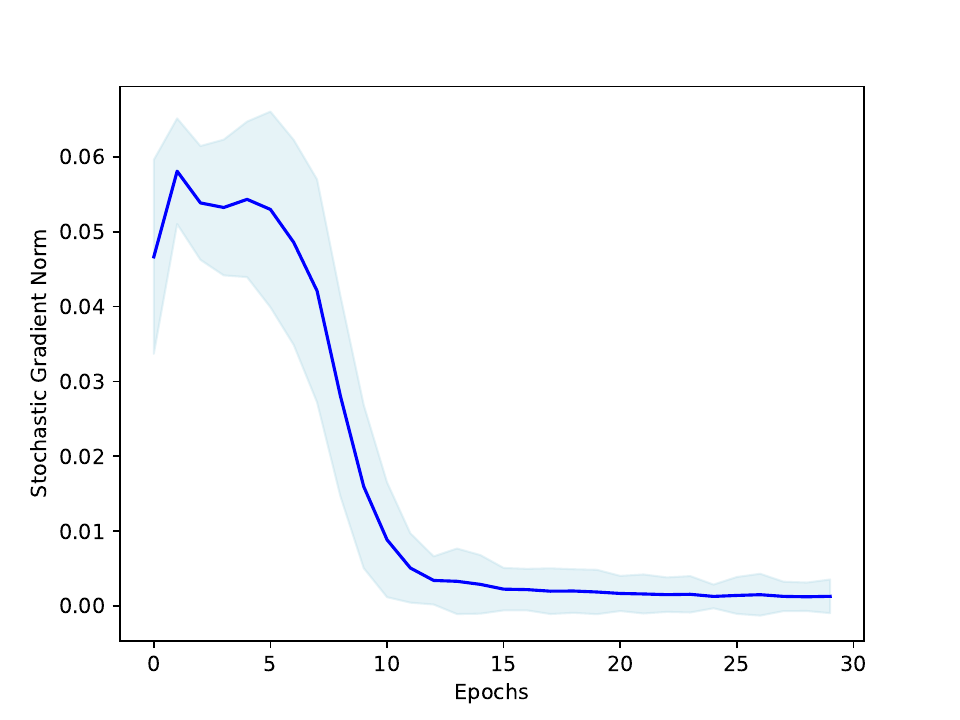}\label{fig:gradient_svhn}}
    \caption{Comparison of gradient norm over three datasets.}
    \label{fig:gradient_norm}
\end{figure}

\subsection{Membership Inference Attack}
Membership Inference Attack (MIA)~\citep{shokri2017membership} is seen as an effective evaluation method of machine unlearning by measuring the privacy leakage of the data supposedly unlearned.
Different from the reproducing verification and the backdoor verification, we tend to categorize MIA into the evaluation method rather than the verification method.
To clarify this, we first aim to distinguish two different settings for measuring the unlearning efficacy. 
We can refer to them as evaluation and verification (which might be mixed up in previous works). 
Evaluation is supposed to be conducted by honest model providers to choose the best unlearning methods from a candidate set of unlearning algorithms. 
In contrast, verification is supposed to be conducted by data owners or third parties to check the unlearning efficacy. 
The most distinct difference between evaluation and verification is the capacity of the evaluator (or verifier), where the evaluator (model provider) has full access to the original model, unlearned model, and the dataset, while the verifier (data owner) only has limited access to the models and the data. 
As a typical example of evaluation methods, comparing the model utility with the retrained model~\citep{golatkar2020eternal,nguyen2022survey} can only be conducted by the model provider with full access.
For MIA, the attacker requires extensive knowledge of the model architecture for white-box attack variants, access to auxiliary or shadow data, and computational power to an extent similar to the model provider~\citep{sommer2022athena}.
However, considering that a powerful third-party verifier can be seen as a potential membership-inference attacker, we conduct additional experiments to demonstrate whether our proposed adversarial methods can circumvent MIA. 
We adapt MIA against machine learning models~\citep{shokri2017membership} to machine unlearning. 
We use a two-layer MLP as the attack model (discriminator). 
The attack model is trained using the same way, aiming at distinguishing the training samples (unlearned samples) from the test samples. 
We label the predictions of the unlearned data as positive data, and we randomly sample predictions of test data to ensure that the number of positive cases and the number of negative cases of the attack model are balanced. 
Ideally, the attack model is supposed to have low accuracy since the unlearned data is supposed to be removed from the trained model and be indistinguishable from the test samples. 
However, if the unlearning is ineffective, the unlearned data is still memorized by the unlearned model as the retained training samples, leading to a high attack accuracy. 
We record the AUC score of the attack under five different random seeds. 
The experimental results are shown in \Cref{tab:membership inference}.
From the results, we can observe that our proposed Adv-R and naive retraining have an attack accuracy around 50\% (similar to random guessing), indicating that our proposed Adv-R is able to circumvent MIA. 
For simple target tasks (e.g. MNIST), the predictions of the test samples are similar to the predictions of the training samples, and they are difficult to distinguish since the model learns well and has good generalizability (the predictions are correct and with high confidence). 
Hence, the attack accuracy is low. 
For difficult target tasks (e.g. CIFAR-10), the model might have a confident and accurate prediction for training samples but not for test samples. 
Subsequently, the training and test samples are easier to distinguish for attack models. 
This can also be seen as a limitation of membership inference attacks (ineffective for well-learned models)~\citep{shafran2021membership}.

\subsection{Effect of Learning Rate on Verification Error}
From \cref{pro:valid_2}, we obtain that we can find a small enough learning rate $\gamma^{(t)}$ and $\gamma_r^{(t)}$ to forge a \textbf{valid} PoRT based on \cref{alg:adv2} for arbitrarily small reproducing error threshold $\varepsilon$.
To verify the effect of learning rate on the verification error, we conduct experiments of different learning rate configurations over the MNIST dataset, using the MLP model.
We choose five different values for learning rate $\gamma^{(t)}$ as $5\times10^{-3}$, $10^{-3}$, $5\times10^{-4}$, $10^{-4}$, and $5\times10^{-5}$. 
To simplify hyperparameter tuning, we directly set the forging learning rate the same as the original learning rate, i.e., $\gamma_r^{(t)}=\gamma^{(t)}$.
Experimental results are shown in \cref{fig:verification_error_lr}. In particular, we show the percentage of verification errors (in different iterations) within the corresponding threshold in \cref{fig:mnist_err}, and show the statistics of verification error and the maximum value (representing the threshold $\varepsilon$ and shown as red triangles) in \cref{fig:mnist_mean_err}.
From the results, we can observe that the verification error threshold (maximum) and the mean value of verification error decreases as the learning rate becomes smaller.
In addition, in the case of $\gamma^{(t)}=\gamma_r^{(t)}$, we can obtain a corollary of \cref{pro:valid_2} as $\varepsilon\geq LC\gamma^2+3C\gamma$ where $C=\max\{L_xC_D,G\}$.
As the value of learning rate $\gamma$ is very small ($\gamma$ usually has a significantly lower order of magnitude compared to $C$), we can approximately omit the second-order term and obtain that $\varepsilon\propto\gamma$, which conforms to the experimental results shown in \cref{fig:mnist_mean_err}.

\subsection{Error Statistics of Reproducing Verification}
We provide an in-depth analysis of the variation trend of the error statistics in \Cref{fig:error_statistics}.
Basically, as mentioned in the experiments, we attribute this result to the change in the gradient norm. 
We first take a look at the verification error from a theoretical perspective. 
Based on the proof of \Cref{pro:valid_2}, when a batch contains the unlearned data, the verification error is related to $b$ (the difference between the gradient on the original data and the gradient on the forging data); when a batch does not contain the unlearned data, the verification error is related to the stochastic gradient (the gradient on a random data point). 
As the number of unlearned data is small (the batches excluding the unlearned data distinctly outnumber the batches including the unlearned data), the mean verification error over an epoch is mainly determined by the norm of the stochastic gradient. 
For MNIST, the model learns well and the optimization converges fast (the accuracy nearly reaches 90\% after the first epoch). 
Hence, the verification error remains stable and is relatively small. For CIFAR-10, the task is challenging for CNN, and the norm of the stochastic gradient grows larger and then decreases as the optimization converges (the convergence is not complete at the end, so we are not able to observe the plateau as MNIST). 
For SVHN, the difficulty of the task is between MNIST and CIFAR-10 for ResNet. Hence, the verification error goes through a short increase and then decrease, and finally goes into a stable plateau as MNIST. 
To support our insights, we plot the norm of the stochastic gradient (we directly use the gradient over the training batch) under different settings. 
The results are shown in \Cref{fig:gradient_norm}. The variation trend of the norm of stochastic gradients matches the verification error shown in \Cref{fig:error_statistics}.
The experimental results demonstrate that the reproducing verification error is closely related to the gradient norm.

\section{Deal with the Vulnerability of MUL Verification}\label{sec:defense}
In the aforementioned studies, we have exposed the vulnerability of the verification strategies of MUL, proved by theoretical and experimental results.
Next, we would like to provide some simple insights on how to deal with vulnerability.

\paragraph{Retraining-based Adversarial Unlearning.} 
Detecting retraining-based adversarial unlearning is extremely difficult because retraining-based adversarial unlearning does not explicitly utilize the unlearned data to update the model parameters.
Consequently, the benefit for model providers from retraining-based adversarial unlearning is relatively small.
In our observation, the performance of the unlearned model conducted by retraining-based adversarial unlearning is better than naive retraining but worse than original training, which conforms to the performance of approximate unlearning methods~\citep{guo2020certified,golatkar2020eternal,golatkar2021mixed,mehta2022deep}.
Unfortunately, existing studies of verification for approximate unlearning remain nascent.
However, it would be promising to investigate the safe verification of approximate unlearning and exploit the verification methods to detect the retraining-based adversarial unlearning method.

\paragraph{Forging-based Adversarial Unlearning.}
Forging-based adversarial unlearning is relatively easy to detect because of its bounded difference from the original model.
According to our experimental results, forging-based adversarial unlearning can be detected by backdoor verification with high confidence (in \cref{tab:backdoor_full}).
Correspondingly, the model provider can largely benefit from forging-based adversarial unlearning (high utility as the original model and low unlearning time cost).
Although theoretically, the model provider can find a proper learning rate $\gamma^{(t)}$ and $\gamma_r^{(t)}$ to circumvent the reproducing verification with arbitrarily small threshold $\varepsilon$, the choice of learning rate is limited in practice.
When model providers choose a smaller learning rate to circumvent more strict reproducing verification, the original training process can take longer time, even fail to converge.
Hence, forging-based adversarial unlearning methods cannot circumvent arbitrarily strict reproducing verification, and the verifier can carefully select the error threshold to reject questionable unlearning operations.


\end{document}